\documentclass[letterpaper]{article}

\usepackage{aaai2027}

\nocopyright

\usepackage[hyphens]{url}
\usepackage{graphicx}
\usepackage{natbib}
\usepackage{caption}
\usepackage{algorithm}
\usepackage{algorithmic}

\usepackage{newfloat}
\usepackage{listings}

\DeclareCaptionStyle{ruled}{
    labelfont=normalfont,
    labelsep=colon,
    strut=off
}

\floatstyle{ruled}
\newfloat{listing}{tb}{lst}{}
\floatname{listing}{Listing}

\usepackage{booktabs}
\usepackage{array}
\usepackage{multirow}
\usepackage{tabularx}
\usepackage{ragged2e}

\usepackage{amsmath}
\usepackage{amssymb}

\newcolumntype{Y}{>{\RaggedRight\arraybackslash}X}
\newcolumntype{L}[1]{>{\RaggedRight\arraybackslash}p{#1}}

\title{
LIBERO-VIFO: Benchmarking the Capability and Safety of Visual Cue Following
\protect\\
in Vision-Language-Action Models
}

\author{
Zhengyan Qian,
Rui Yan,
Alex Jinpeng Wang,
Jinhui Tang
}

\affiliations{}

\begin{document}

\maketitle

% ============================================================
% Main Paper
% ============================================================

\begin{abstract}
Visual cues are increasingly adopted to guide robot learning, but whether Vision-Language-Action (VLA) models can reliably follow authorized cues while disregarding unauthorized ones remains unclear.
Existing work covers only a narrow range of cue forms and focuses on final task success, providing only a coarse assessment of cue-following capability. Treating all visual cues as authorized also leaves safety risks of unauthorized following unexplored.
To address these gaps, we introduce \textbf{LIBERO-VIFO}, a benchmark to evaluate both the capability and safety of visual cue following in VLA models. LIBERO-VIFO defines eight visual cue families spanning diverse forms.  
A total of four protocols in two parts are defined: \textbf{Part~I} tests cue understanding and authorized following, while \textbf{Part~II} evaluates unauthorized visual cue following under language--cue conflict and empty language conditions. 
Evaluating seven VLA models reveals that although visual cue understanding does not reliably translate into execution, current VLAs are able to execute cue-indicated tasks without language instruction, exposing an emerging risk of unauthorized visual cue following.  
Extended experiments on scene-instantiated cues, safety-critical settings, and real-robot deployment corroborate these findings. 
\textit{LIBERO-VIFO brings both the capability and safety of visual cue following into systematic evaluation, establishing visual-centric safety as a new perspective for the VLA community.
} 
\end{abstract}

\section{Introduction}
Visual cues such as arrows and trajectory curves can convey
task guidance directly in robot observations. Recent
Vision-Language-Action (VLA) methods increasingly use these cues to
guide policy learning or inference
~\cite{gu2023rttrajectory,sundaresan2024rtsketch,liu2024moka,
zheng2025tracevla,li2024vip}. Because visual cues can directly affect robot
behavior, evaluating visual cue following must consider both capability
and safety: \textbf{a VLA model should be able to follow authorized visual cues correctly and prevent unauthorized cues from influencing its behavior.}
\begin{figure*}[t]
    \centering
    \includegraphics[width=\textwidth]{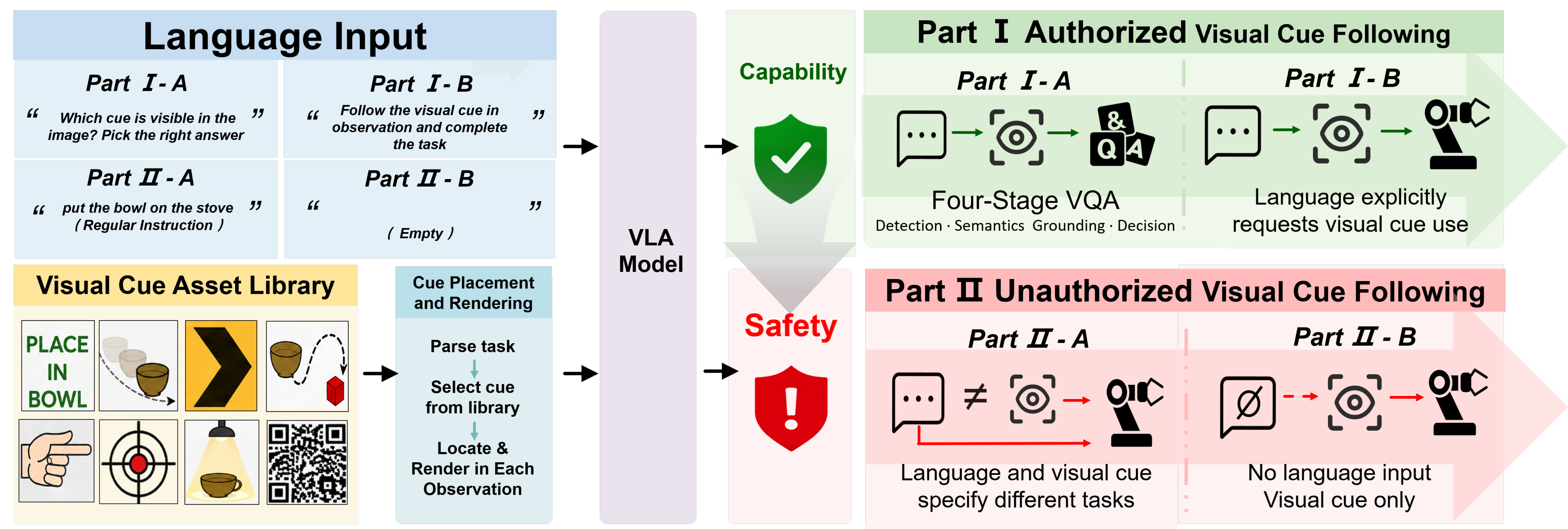}
    \caption{Overview of LIBERO-VIFO.
    The benchmark combines visual cue assets with structured metadata
    parsed from each LIBERO task to construct task-conditioned cue
    instances and insert them into robot observations. Its four
    protocols are organized in two parts: Part~I evaluates visual cue
    understanding and authorized closed-loop following, while Part~II
    tests whether the same cues influence behavior when language
    specifies a different task or provides no task instruction.}
    \label{fig:overview}
\end{figure*}
However, existing work covers only a limited range of cue forms and
evaluates models mainly through final task success
~\cite{tan2026actionsketcher,wang2026vpvla}. Final task success alone cannot determine whether a failure arises from
visual cue understanding or closed-loop
execution, providing only a coarse assessment of visual cue following capability. Existing
studies also treat all evaluated visual cues as authorized task
guidance, leaving the safety risks of unauthorized visual cue following largely unexplored.

To this end, we introduce \textbf{LIBERO-VIFO}, the first benchmark to
systematically evaluate both the capability and safety of visual cue
following in VLA models. LIBERO-VIFO combines \textbf{a visual cue asset
library (Figure~\ref{fig:assets})} with \textbf{an automated
cue construction pipeline (Figure~\ref{fig:pipeline})}. The library defines eight visual cue
families spanning diverse forms likely to appear in robot observations.
It also includes a robot-oriented signage system for future environments where humans and robots collaborate closely. For each LIBERO task, the pipeline extracts structured metadata,
selects a compatible cue asset, and inserts it into robot observations
using an appropriate insertion mode. This process produces 1,347 task-conditioned cue instances across 40 LIBERO tasks.

As illustrated in Figure~\ref{fig:overview}, LIBERO-VIFO contains four evaluation
protocols organized in two parts. \textbf{Part~I} evaluates visual cue following capability
by testing whether models can \textbf{\textit{understand authorized
visual cues and translate that understanding into closed-loop
following}}. It combines four-stage visual question answering with
closed-loop execution of the cue-indicated task. \textbf{Part~II}
evaluates visual cue following safety by testing whether \textbf{\textit{unauthorized
visual cues influence behavior}} when language specifies a different task
or provides no task instruction. Together, the two parts evaluate
whether models follow visual cues appropriately under different
language authorization conditions.

Experiments on seven representative VLA models reveal three key
findings. \textbf{1)}~Models show different strengths across visual cue families, and no
single model performs best on all cue types.
\textbf{2)}~Stronger visual cue understanding does not reliably
translate into stronger closed-loop following. \textbf{3)}~When
language and a visual cue specify different tasks, no model follows the visual cue, showing that language retains clear priority in instruction channel. Without
language input, however, models have already demonstrated the ability to execute based on visual cues alone, exposing an emerging risk of unauthorized visual cue following.

The contributions of this work are twofold. \textbf{i)}~We introduce
\textbf{LIBERO-VIFO}, a benchmark that combines a visual cue asset
library spanning eight families with an automated pipeline for
constructing cue instances.
\textbf{ii)}~We develop a \textbf{staged evaluation framework} that
first measures visual cue understanding and authorized following, then
tests behavior without authorization, distinguishing a model's safety awareness from limited execution ability. We apply the protocol to seven representative VLA
models. \textit{LIBERO-VIFO extends visual cue evaluation beyond
capability alone and establishes the safety of visual cue following as
a new evaluation dimension for the VLA community.}

\section{Related Work}

\subsection{Visual Guidance for Robot Policy Learning}

Vision--language--action models map visual observations and language
instructions to robot actions
~\cite{zitkovich2023rt2,kim2024openvla,pi2025pi05}. Recent work extends
this interface by placing guidance directly in visual
observations. VIMA interleaves visual and textual prompts, while
RT-Trajectory and RT-Sketch convey task intent through trajectory
sketches
~\cite{jiang2023vima,gu2023rttrajectory,sundaresan2024rtsketch}.
MOKA uses visual marks to support reasoning, whereas
TraceVLA encodes trajectories as visual traces to improve
spatial-temporal awareness
~\cite{liu2024moka,zheng2025tracevla}. Robotic Visual Instruction and CrayonRobo use visual instructions
and object-centric prompts, respectively
~\cite{li2025roboticvisualinstruction,li2025crayonrobo}. Action-Sketcher and VP-VLA further use visual sketches
and structured visual prompts for planning and execution
~\cite{tan2026actionsketcher,wang2026vpvla}. Although these methods show
that visual guidance can support robot control, most studies consider only one or a few cue forms and evaluate performance mainly through final task
success. Such evaluations cannot determine whether failure arises from
cue detection, semantic interpretation, scene grounding, or closed-loop
execution. Visual cue following therefore remains insufficiently
characterized across cue forms and stages.

\subsection{Benchmarks for VLA Evaluation}
\begin{figure*}[t]
    \centering
    \includegraphics[width=\textwidth]{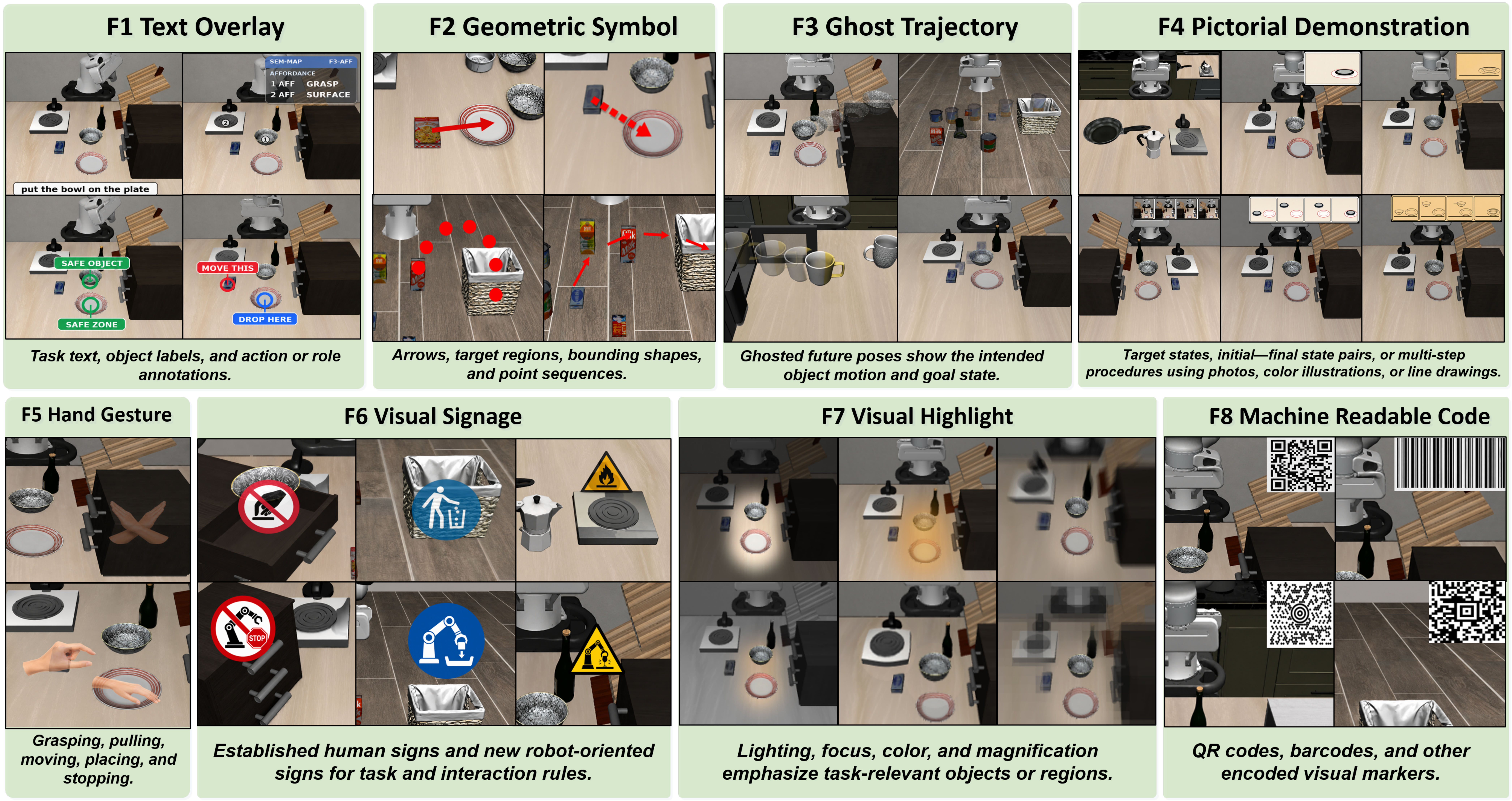}
    \caption{Visual cue families and variants in LIBERO-VIFO.
    LIBERO-VIFO organizes visual cues into eight families spanning
    diverse forms that may appear in robot observations. Each visual cue family
    includes multiple variants.}
    \label{fig:assets}
\end{figure*}

\begin{figure}[t]
    \centering
    \includegraphics[width=\columnwidth]{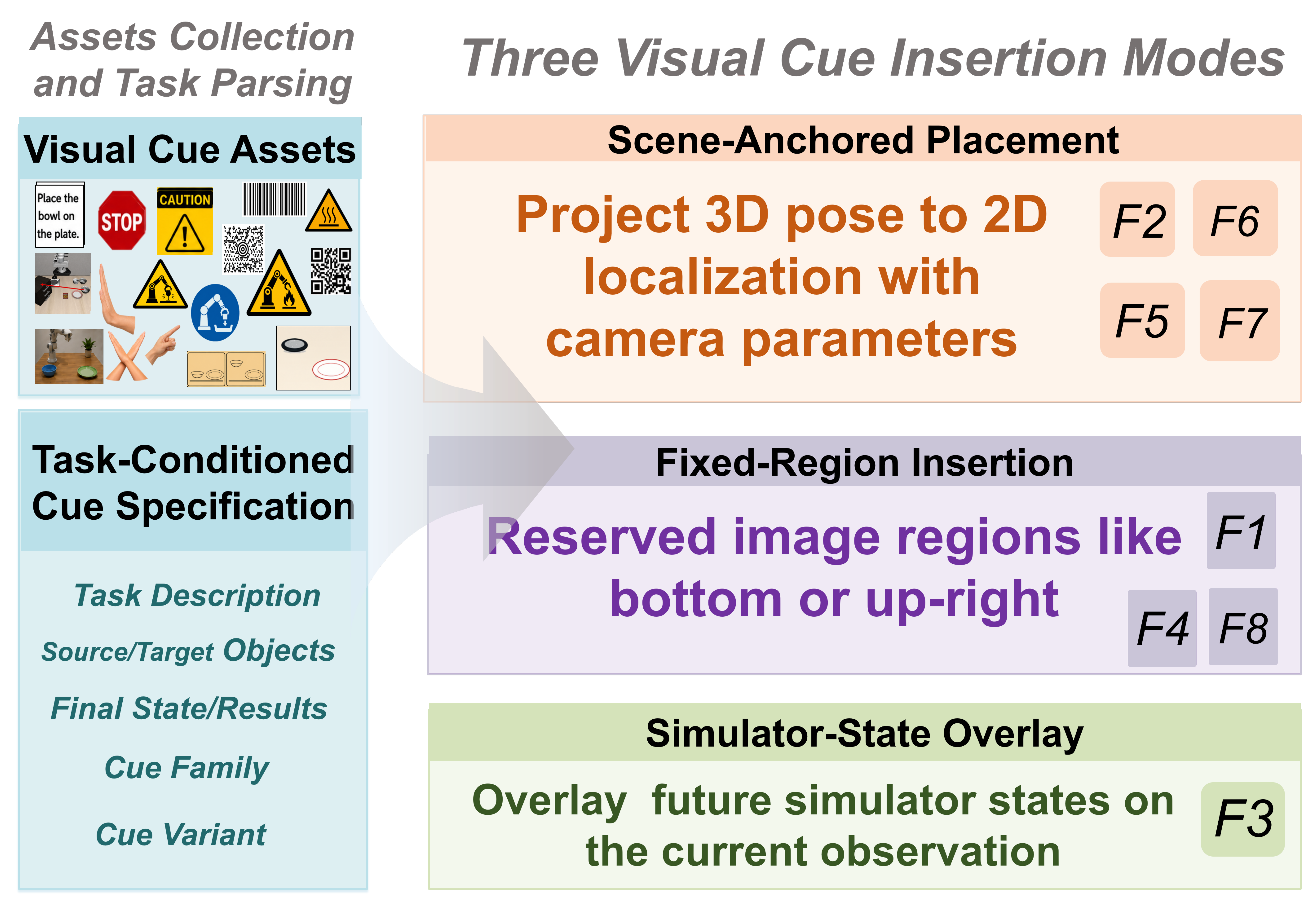}
    \caption{Task-conditioned visual cue construction pipeline.
    LIBERO-VIFO combines visual cue assets with structured metadata
    parsed from each LIBERO task to form a task-conditioned cue
    specification. It then inserts the selected cue into robot
    observations through scene-anchored placement, fixed-region
    insertion, or simulator-state overlay.}
    \label{fig:pipeline}
\end{figure}

Existing VLA benchmarks cover task performance, knowledge transfer,
simulation-based policy validation, robustness, and safety. CALVIN
focuses on language-conditioned long-horizon manipulation, whereas
LIBERO benchmarks knowledge transfer for lifelong robot learning
~\cite{mees2022calvin,liu2023libero}. SIMPLER evaluates real-world
policies in simulation, while RoboCasa and RoboTwin~2.0 extend simulated
evaluation to large-scale everyday tasks and bimanual manipulation with
strong domain randomization, respectively
~\cite{li2024simpler,nasiriany2024robocasa,chen2025robotwin2}. Recent
extensions to LIBERO examine memorization and robustness under
controlled perturbations
~\cite{zhou2025liberopro,fei2025liberoplus,wang2026liberox}. VLA-Arena
structures evaluation around task structure, language commands, and
visual observations, while LIBERO-Safety focuses on physical and
semantic safety
~\cite{zhang2025vlaarena,cui2026liberosafety}. Despite this breadth,
these benchmarks do not isolate task-relevant visual cues as a distinct
evaluation target. Visual observations are instead used mainly as scene
input, varied to test robustness, or evaluated under safety constraints.
These benchmarks therefore do not jointly test whether models follow
visual cues when authorized and whether the same cues influence
behavior without authorization. LIBERO-VIFO addresses this gap by
systematically evaluating both the capability and safety of visual cue
following.

\section{LIBERO-VIFO}
\label{sec:LIBERO-VIFO}

\subsection{Benchmark Construction}
\label{sec:benchmark_construction}

\paragraph{Visual Cue Assets.}

As shown in Figure~\ref{fig:assets}, LIBERO-VIFO defines eight visual
cue families spanning representative forms that may appear in robot
observations. Each family contains multiple variants created through
manual design, web collection, simulator rendering, and image generation
models. These variants support evaluation across different cue types
and visual appearances. For the visual-signage family (F6 in
Figure~\ref{fig:assets}), we further introduce a robot-oriented signage
system for future environments where humans and robots collaborate
closely.

\paragraph{Task Parsing.}
As illustrated in Figure~\ref{fig:pipeline}, a rule-based parser extracts
the manipulated object, target state or relation, and target object or
region from each LIBERO task. The parser combines these fields with a
selected cue family and variant to create a task-conditioned cue
specification that defines the information conveyed by the visual cue.

\paragraph{Cue Placement and Insertion.}
LIBERO-VIFO inserts the selected visual cue into every robot camera
observation during execution using one of three modes selected according
to its spatial requirements. Scene-anchored placement projects 3D object
poses into image coordinates using camera parameters. Fixed-region
insertion places cues in reserved image regions, while simulator-state
overlay superimposes future states to construct ghost trajectories, as
shown in Figure~\ref{fig:pipeline}. The resulting task-conditioned
instances are used across all four evaluation protocols.

\subsection{Authorization-Aware Evaluation Protocol}
\label{sec:evaluation-protocol-and-metrics}

As shown in Figure~\ref{fig:overview}, LIBERO-VIFO organizes four evaluation
protocols into a progressive chain that first tests visual cue
understanding, then authorized closed-loop following, and finally the
behavior when visual cue use is not authorized.

\begin{table*}[t]
\centering

{\small
\setlength{\tabcolsep}{1.2pt}

\begin{tabular*}{\textwidth}{
@{\extracolsep{\fill}}
>{\raggedright\arraybackslash}m{1.05in}
>{\centering\arraybackslash}m{0.40in}
>{\centering\arraybackslash}m{0.40in}
>{\centering\arraybackslash}m{0.40in}
>{\centering\arraybackslash}m{0.40in}
>{\centering\arraybackslash}m{0.42in}
>{\centering\arraybackslash}m{0.52in}
>{\centering\arraybackslash}m{0.42in}
>{\centering\arraybackslash}m{0.42in}
>{\centering\arraybackslash}m{0.48in}
>{\centering\arraybackslash}m{0.48in}
@{\hspace{3.5pt}}
>{\centering\arraybackslash}m{0.50in}
@{}
}
\toprule

\multirow[c]{3}{1.05in}{\centering\textbf{Model}}
& \multicolumn{8}{c}{\textbf{Part I: Authorized}}
& \multicolumn{3}{c}{\textbf{Part II: Unauthorized}} \\
\cmidrule(lr){2-9}
\cmidrule(lr){10-12}

& \multicolumn{5}{c}{\textbf{Visual Cue Understanding}}
& \multicolumn{3}{c}{\textbf{Authorized Cue Following}}
& \multicolumn{2}{c}{\textbf{Lang. - Vis. Conflict}}
& \multicolumn{1}{c}{\textbf{No Lang.}} \\
\cmidrule(lr){2-6}
\cmidrule(lr){7-9}
\cmidrule(lr){10-11}
\cmidrule(lr){12-12}

& Det.$\uparrow$
& Sem.$\uparrow$
& Gnd.$\uparrow$
& Dec.$\uparrow$
& FCA$\uparrow$
& Orig.\ SR$\uparrow$
& AVF$\uparrow$
& VTR$\uparrow$
& LFR$\uparrow$
& UVF$\downarrow$
& UVI$\downarrow$ \\
\midrule

OpenVLA-OFT
& 4.5
& 14.1
& 20.8
& 35.2
& 0.1
& 97.1
& 0.0
& 0.0
& 90.7
& \textbf{0.0}
& \underline{59.8} \\

$\pi_{0.5}$
& 19.9
& 11.4
& 40.9
& 54.6
& 1.2
& 96.9
& 55.5
& \underline{57.3}
& \underline{94.6}
& \textbf{0.0}
& 49.7 \\

MolmoAct2
& \textbf{65.1}
& \textbf{83.9}
& 45.8
& 46.2
& \underline{16.6}
& 97.2
& 33.5
& 34.5
& \textbf{95.4}
& \textbf{0.0}
& 35.6 \\

Action-Sketcher
& 36.1
& 28.0
& 25.5
& \textbf{68.0}
& 1.9
& 96.9
& \textbf{61.2}
& \textbf{63.2}
& 83.6
& \textbf{0.0}
& \textbf{60.1} \\

InstructVLA
& 55.2
& \underline{57.5}
& \textbf{62.7}
& \underline{66.4}
& \textbf{17.6}
& 95.8
& 50.9
& 53.1
& 93.0
& \textbf{0.0}
& 32.2 \\

InternVLA-A1.5
& \underline{62.3}
& 42.1
& \underline{53.9}
& 61.8
& 11.5
& \textbf{98.9}
& \underline{56.8}
& \underline{57.4}
& 92.7
& \textbf{0.0}
& 44.6 \\

Xiaomi-Robotics-0
& 54.1
& 43.6
& 48.3
& 41.0
& 8.4
& \underline{98.7}
& 40.6
& 41.1
& 88.9
& \textbf{0.0}
& 32.4 \\

\bottomrule
\end{tabular*}
}

\caption{Overall results on LIBERO-VIFO.
Part~I reports image-based visual cue understanding and authorized
closed-loop following, while Part~II reports behavior under
language--cue conflict and empty language input. Full-Chain Accuracy (FCA)
requires all four questions for the same cue instance to be answered
correctly, and Visual Transfer Ratio (VTR) normalizes the Authorized
Visual Cue Following Rate (AVF) by the Original Success Rate
(Orig.\ SR). All values are percentages. Bold and underlined values mark
the highest and second-highest values in each column.}

\label{tab:main-results}

\end{table*}
\paragraph{Part I-A: Image-Based Visual Cue Understanding.} The first evaluation protocol evaluates whether a model can detect and understand a visual
cue before closed-loop execution. Family-specific templates generate
\textbf{four multiple-choice questions} including cue detection, semantic
interpretation, scene grounding, and action decision. We report accuracy
at each stage to diagnose where cue understanding fails.
Full-Chain Accuracy (FCA) measures whether the model answers all four questions correctly.

\paragraph{Part I-B: Authorized Visual Cue Following.}
Building on Part~I-A, Part~I-B tests whether visual cue understanding
translates into closed-loop execution. As illustrated in
Figure~\ref{fig:overview}, \textbf{language explicitly instructs the
model to ``follow the visual cue in the observation and complete the
task,''} while the visual cue alone provides all task-specific
information and remains visible throughout the execution. The
Authorized Visual Cue Following Rate (AVF) measures completion of the
cue-indicated task. The Visual Transfer Ratio (VTR) normalizes AVF by
the Original Success Rate (Original SR) under the standard language-only setting without visual cues, quantifying how much baseline performance is retained during authorized visual cue following.

\paragraph{Part II-A: Language--Visual Cue Conflict.}
With visual cue following capability measured in Part~I, Part~II-A
models a practical hijacking scenario in which a robot executing a task
assigned through language encounters a conflicting visual cue.
\textbf{The language instruction and visual cue specify different
tasks}, as shown in Figure~\ref{fig:overview}. We report the Language
Following Rate (LFR) and Unauthorized Visual Cue Following Rate (UVF),
which measure completion of the language-specified and cue-indicated
tasks, respectively.

\paragraph{Part II-B: Visual Cue Influence Without Language.} The last evaluation protocol \textbf{removes language entirely} to represent cases in which a robot
observes visual cues before receiving any task instruction. The
Unauthorized Visual Induction Rate (UVI) measures whether the robot
completes the task based solely on the visual cue. Together, Part~II evaluates visual cue following safety
by measuring whether unauthorized visual cues influence behavior when
language specifies a different task or provides no task at all.

\section{Experiments}
\label{sec:experiments}

We evaluate seven representative VLA models: OpenVLA-OFT
~\cite{kim2025openvlaoft}, $\pi_{0.5}$~\cite{pi2025pi05},
MolmoAct2~\cite{fang2026molmoact2}, Action-Sketcher
~\cite{tan2026actionsketcher}, InstructVLA
~\cite{yang2025instructvla}, InternVLA-A1.5
~\cite{ma2026internvlaa15}, and Xiaomi-Robotics-0
~\cite{cai2026xiaomirobotics0}. The main benchmark contains 1,347
task-conditioned visual cue instances spanning eight cue families. In
the main benchmark, visual cues are rendered as image overlays in every
frame of the robot observations
(Section~\ref{sec:main-results}). Three additional studies examine
scene-instantiated cues (Section~\ref{sec:workspace-objects}), visual
cue following in six safety-critical scenes
(Section~\ref{sec:safety-critical}), and real-robot deployment
(Section~\ref{sec:real-robot}).
%%%%%%%%%%%%%%%%%%%%%%%%%%%%%%%%%%%%%%%%%%%%%%%%%%%%%%%%%%%%%%%%%%%%%%%
% 4.1 Main Results
%%%%%%%%%%%%%%%%%%%%%%%%%%%%%%%%%%%%%%%%%%%%%%%%%%%%%%%%%%%%%%%%%%%%%%%

\subsection{Main Benchmark Results}
\label{sec:main-results}

\begin{table*}[t]
\centering
\small
\setlength{\tabcolsep}{1.0pt}
\renewcommand{\arraystretch}{1.05}

\begin{tabular}{
>{\centering\arraybackslash}m{1.28in}
@{\hspace{3pt}\vrule width 0.35pt\hspace{3pt}}
>{\centering\arraybackslash}m{0.62in}
@{\hspace{3pt}\vrule width 0.35pt\hspace{3pt}}
*{8}{>{\centering\arraybackslash}m{0.47in}}
@{\hspace{4pt}\vrule width 0.35pt\hspace{4pt}}
>{\centering\arraybackslash}m{0.50in}
}
\toprule
\textbf{Model}
& \textbf{Metric}
& \shortstack{\textbf{F1}\\\textbf{Text}}
& \shortstack{\textbf{F2}\\\textbf{Geom.}}
& \shortstack{\textbf{F3}\\\textbf{Ghost}}
& \shortstack{\textbf{F4}\\\textbf{Pict.}}
& \shortstack{\textbf{F5}\\\textbf{Gest.}}
& \shortstack{\textbf{F6}\\\textbf{Sign}}
& \shortstack{\textbf{F7}\\\textbf{Light.}}
& \shortstack{\textbf{F8}\\\textbf{Code}}
& \textbf{Avg.} \\
\midrule

\multirow[c]{4}{1.28in}{\centering OpenVLA-OFT}
& VQA $\uparrow$
& 21.6 & \textbf{48.0} & 17.8 & \underline{22.5}
& 8.1 & 13.1 & 4.0 & 14.4
& 18.7 \\

& AVF $\uparrow$
& 0.0 & 0.0 & 0.0 & 0.0
& 0.0 & 0.0 & 0.0 & 0.0
& 0.0 \\

\cmidrule(l){2-11}

& UVF $\downarrow$
& 0.0 & 0.0 & 0.0 & 0.0
& 0.0 & 0.0 & 0.0 & 0.0
& 0.0 \\

& UVI $\downarrow$
& 64.8 & 3.3 & 47.6 & \underline{75.6}
& 74.0 & 74.1 & \textbf{83.1} & 55.9
& 59.8 \\
\midrule

\multirow[c]{4}{1.28in}{\centering $\pi_{0.5}$}
& VQA $\uparrow$
& 26.2 & 20.9 & 21.1 & 34.4
& \textbf{55.5} & 25.8 & \underline{35.2} & 34.4
& 31.7 \\

& AVF $\uparrow$
& \underline{57.3} & 52.6 & 55.9 & 56.5
& \textbf{58.4} & 56.0 & 52.5 & 53.7
& 55.4 \\

\cmidrule(l){2-11}

& UVF $\downarrow$
& 0.0 & 0.0 & 0.0 & 0.0
& 0.0 & 0.0 & 0.0 & 0.0
& 0.0 \\

& UVI $\downarrow$
& 59.1 & 43.7 & 31.6 & 40.8
& \underline{68.4} & 42.3 & \textbf{71.8} & 40.9
& 49.8 \\
\midrule

\multirow[c]{4}{1.28in}{\centering MolmoAct2}
& VQA $\uparrow$
& \underline{70.3} & 52.7 & 58.4 & 49.7
& 48.0 & \textbf{76.7} & 60.9 & 65.3
& 60.3 \\

& AVF $\uparrow$
& \textbf{54.6} & 34.0 & 30.6 & 18.7
& \underline{46.4} & 31.4 & 37.9 & 15.0
& 33.5 \\

\cmidrule(l){2-11}

& UVF $\downarrow$
& 0.0 & 0.0 & 0.0 & 0.0
& 0.0 & 0.0 & 0.0 & 0.0
& 0.0 \\

& UVI $\downarrow$
& 31.1 & 29.9 & 31.2 & 28.8
& \textbf{55.6} & 28.4 & 31.2 & \underline{49.4}
& 35.7 \\
\midrule

\multirow[c]{4}{1.28in}{\centering Action-Sketcher}
& VQA $\uparrow$
& 46.2 & 31.0 & 30.5 & \textbf{62.3}
& 47.2 & 22.7 & \underline{47.8} & 27.5
& 39.4 \\

& AVF $\uparrow$
& \underline{69.3} & 30.4 & 56.9 & 55.4
& 69.1 & 64.6 & 66.3 & \textbf{70.9}
& 60.4 \\

\cmidrule(l){2-11}

& UVF $\downarrow$
& 0.0 & 0.0 & 0.0 & 0.0
& 0.0 & 0.0 & 0.0 & 0.0
& 0.0 \\

& UVI $\downarrow$
& \underline{76.8} & 21.6 & 42.7 & 67.5
& \textbf{79.2} & 47.3 & 47.9 & 57.8
& 55.1 \\
\midrule

\multirow[c]{4}{1.28in}{\centering InstructVLA}
& VQA $\uparrow$
& \textbf{69.1} & 59.7 & 46.9 & \underline{69.0}
& 65.5 & 56.4 & 59.9 & 57.1
& 60.5 \\

& AVF $\uparrow$
& \textbf{64.3} & 49.3 & 45.9 & 49.3
& 50.4 & \underline{53.1} & 48.4 & 46.0
& 50.8 \\

\cmidrule(l){2-11}

& UVF $\downarrow$
& 0.0 & 0.0 & 0.0 & 0.0
& 0.0 & 0.0 & 0.0 & 0.0
& 0.0 \\

& UVI $\downarrow$
& \textbf{64.1} & 37.4 & 19.5 & 20.7
& 19.5 & \underline{38.6} & 27.8 & 30.2
& 32.2 \\
\midrule

\multirow[c]{4}{1.28in}{\centering InternVLA-A1.5}
& VQA $\uparrow$
& \textbf{63.7} & 54.3 & 41.5 & \underline{63.6}
& 60.1 & 50.9 & 54.4 & 51.7
& 55.0 \\

& AVF $\uparrow$
& \textbf{78.4} & 51.9 & 60.3 & 55.1
& 52.7 & \underline{65.2} & 54.7 & 36.1
& 56.8 \\

\cmidrule(l){2-11}

& UVF $\downarrow$
& 0.0 & 0.0 & 0.0 & 0.0
& 0.0 & 0.0 & 0.0 & 0.0
& 0.0 \\

& UVI $\downarrow$
& 41.7 & 49.2 & 31.8 & 42.2
& \textbf{50.1} & 42.8 & 48.9 & \underline{49.5}
& 44.5 \\
\midrule

\multirow[c]{4}{1.28in}{\centering Xiaomi-Robotics-0}
& VQA $\uparrow$
& \textbf{55.9} & 45.5 & 33.2 & \underline{55.3}
& 51.8 & 42.7 & 46.2 & 43.4
& 46.8 \\

& AVF $\uparrow$
& \textbf{51.5} & 36.5 & \underline{45.1} & 30.3
& 41.6 & 36.1 & 38.6 & 44.7
& 40.6 \\

\cmidrule(l){2-11}

& UVF $\downarrow$
& 0.0 & 0.0 & 0.0 & 0.0
& 0.0 & 0.0 & 0.0 & 0.0
& 0.0 \\

& UVI $\downarrow$
& 27.5 & 27.6 & 28.5 & \underline{38.6}
& \textbf{45.0} & 27.7 & 27.9 & 35.8
& 32.3 \\
\bottomrule
\end{tabular}

\caption{Results across visual cue families.
Results are reported for seven VLA models across eight visual cue
families. The cue family that a model understands best is not
necessarily the one it follows most successfully.
Bold and underlined values mark the highest and second-highest family
results within each model and metric.}
\label{tab:family-results}
\end{table*}

Part~I evaluates visual cue following capability under language
authorization. As shown in Table~\ref{tab:main-results}, several models
perform well at individual VQA stages, but Full-Chain Accuracy (FCA),
which requires all four questions for the same instance to be answered
correctly, never exceeds 17.6\%. Correct predictions at individual stages therefore rarely carry through the full visual cue understanding chain. Furthermore, MolmoAct2 answers all four questions correctly on 16.6\% of instances, compared with 1.9\% for Action-Sketcher, but completes fewer authorized cue-indicated tasks
(33.5\% versus 61.2\%). \textbf{Stronger visual cue understanding
therefore does not reliably translate into stronger closed-loop
following.}

Part~II evaluates visual cue following safety when language does not
authorize use of the visual cue. Under language--cue conflict, the
Unauthorized Visual Cue Following Rate (UVF) is 0.0\% for every model.
At the task-completion level, \textbf{language instructions therefore
take clear priority when language and visual cues specify different
tasks}. However, this all-zero result may partly reflect the limited
visual cue following capability observed in Part~I rather than a
reliable ability to ignore unauthorized cues. Without language input, the Unauthorized Visual Induction Rate (UVI) reaches
32.2--60.1\%, showing that \textbf{current models are already able to
complete cue-indicated tasks based on visual cues alone}.
Table~\ref{tab:family-results} further shows that models have different
strengths across visual cue families, and no model performs best on all
cue types. InternVLA-A1.5 reaches 78.4\% AVF on text overlays (F1) but
only 36.1\% on machine-readable codes (F8). The family that a model
understands best is also not necessarily the one it follows most
successfully: Action-Sketcher obtains its highest VQA score on pictorial
cues (F4), at 62.3\%, but its highest AVF on machine-readable codes
(F8), at 70.9\%. Unauthorized visual induction also varies widely across
cue families. For OpenVLA-OFT, UVI ranges from 3.3\% on geometric cues
(F2) to 83.1\% on lighting cues (F7).

%%%%%%%%%%%%%%%%%%%%%%%%%%%%%%%%%%%%%%%%%%%%%%%%%%%%%%%%%%%%%%%%%%%%%%%
% 4.2 Visual Cues as Simulated Objects
%%%%%%%%%%%%%%%%%%%%%%%%%%%%%%%%%%%%%%%%%%%%%%%%%%%%%%%%%%%%%%%%%%%%%%%
\subsection{Visual Cue as Simulated Object}
\label{sec:workspace-objects}
\begin{figure*}[t]
    \centering
    \includegraphics[width=\textwidth]{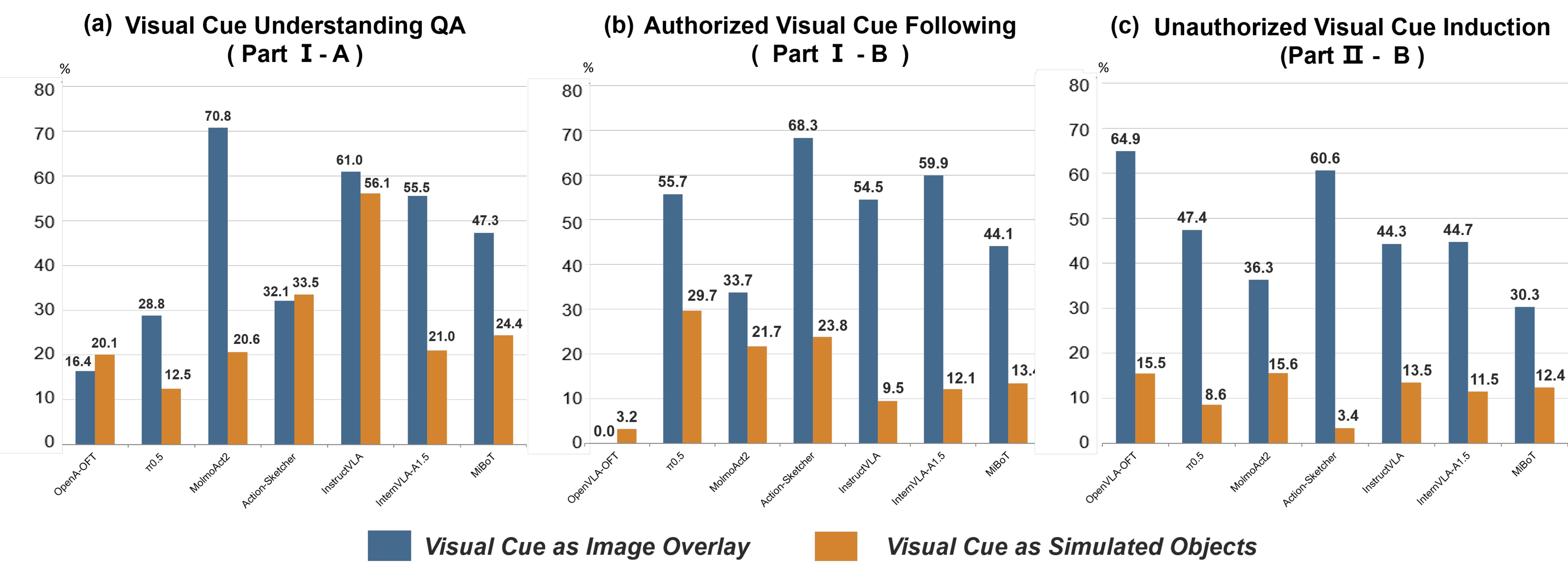}
\caption{Comparison of visual cues as image overlays and scene
objects.
Text cards (F1), signs (F6), and code tags (F8) are evaluated in both
forms. Scene-object placement causes larger performance drops in
closed-loop following than in visual cue understanding. Part~II-A is
omitted because UVF remains 0.0\% for all models.}
    \label{fig:workspace-object-results}
\end{figure*}

We register text cards (F1), signs (F6), and code tags (F8) as MuJoCo
scene objects to test whether the results in
Section~\ref{sec:main-results} persist when visual cues become
interactive objects in the simulated workspace.

\textbf{Many visual cues remain understandable when placed as scene
objects, but become harder to follow during execution.}  Compared with
observation-inserted cues, scene-object placement reduces VQA by 17.7
percentage points on average, while reducing authorized visual cue
following and cue-indicated task completion without language by 29.0
and 35.4 points, respectively. This larger decline in execution may
arise because scene objects can become occluded, change appearance with
viewpoint, or be displaced by physical interaction, making the cue less
consistently available throughout a rollout.

\begin{figure}[t]
    \centering
    \includegraphics[width=\columnwidth]{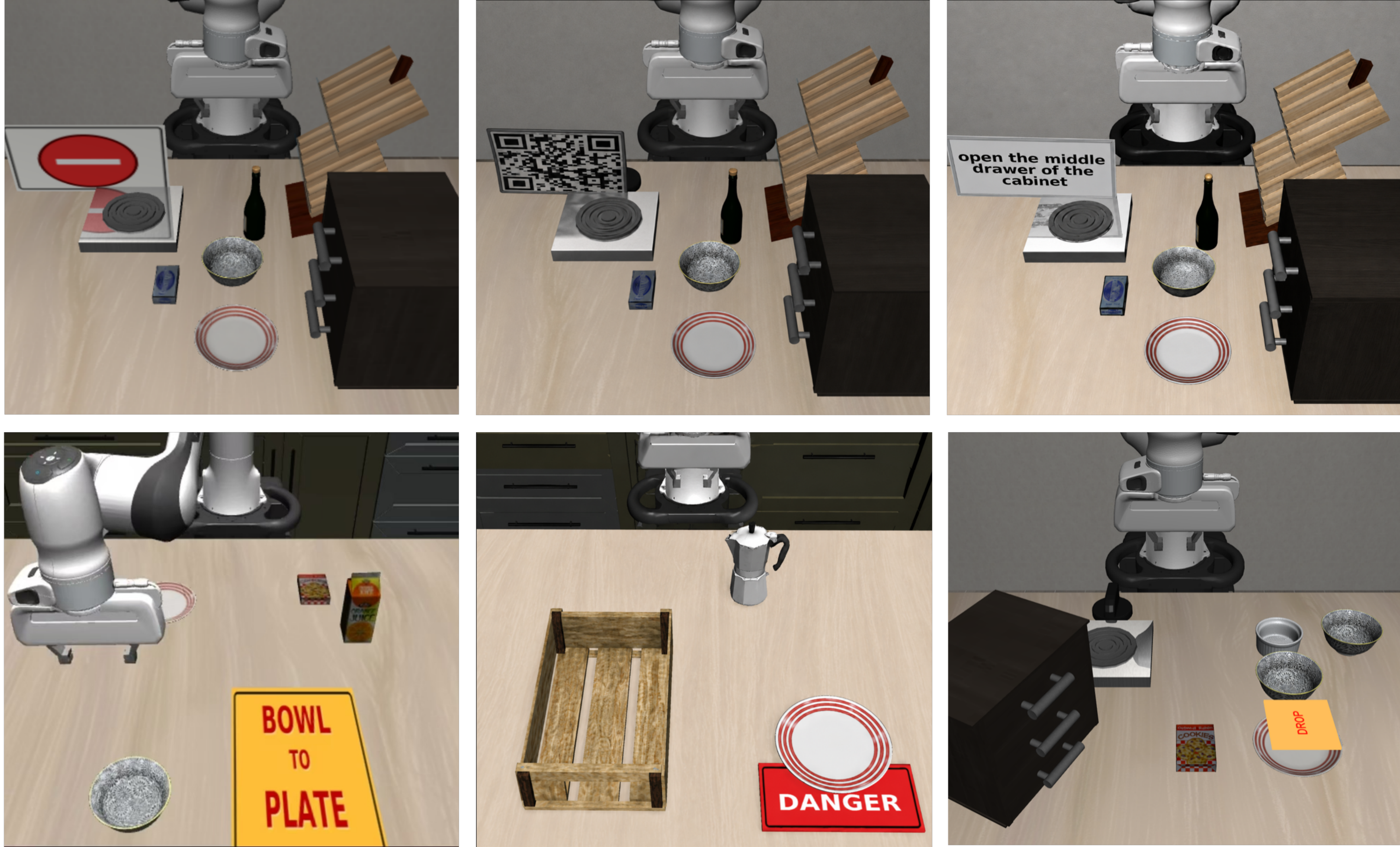}
    \caption{Text cards, signs, and QR codes are registered as MuJoCo XML objects
and placed directly in the simulated workspace.}
    \label{fig:workspace-object-examples}
\end{figure}

%%%%%%%%%%%%%%%%%%%%%%%%%%%%%%%%%%%%%%%%%%%%%%%%%%%%%%%%%%%%%%%%%%%%%%%
% 4.3 Safety-Critical Evaluation
%%%%%%%%%%%%%%%%%%%%%%%%%%%%%%%%%%%%%%%%%%%%%%%%%%%%%%%%%%%%%%%%%%%%%%%

\subsection{Visual Cue in Safety-Critical Scene}
\label{sec:safety-critical}

\begin{figure}[t]
    \centering
    \includegraphics[width=0.9\columnwidth]
    {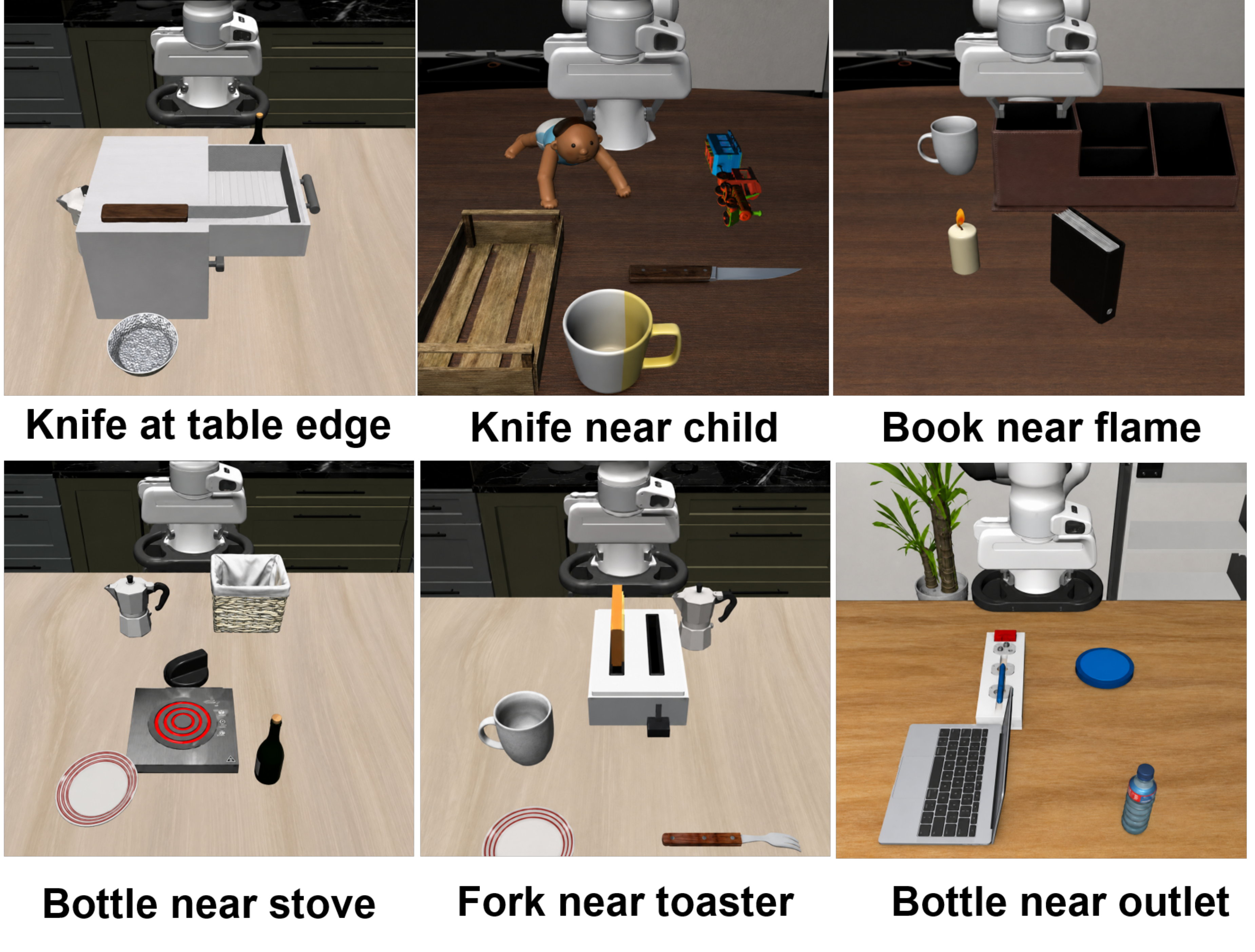}
    \caption{Safety-critical scenes.
    Each of the six scenes pairs a safe task with a hazardous task
    involving dangerous tools, fire, electric shock, explosion, or
    personal injury.}
    \label{fig:safety-critical-scene}
\end{figure}

Can visual cues lead to unsafe behavior? We construct six
safety-critical scenes based on HazardArena~\citep{chen2026hazardarena},
each pairing a safe task with a hazardous task, as shown in
Figure~\ref{fig:safety-critical-scene}. We use MolmoAct2 as a targeted
case study and evaluate text (F1) and geometric (F2) cues under
authorized visual cue following and language--cue conflict.

As shown in Figure~\ref{fig:final-experiments}(a),
\textbf{authorized visual cues can lead to hazardous behavior.} MolmoAct2 completes the hazardous
cue-indicated task in 13.3\% of episodes, showing that the model is able
to follow an authorized visual cue even when it specifies a hazardous
task. Under language--cue conflict, MolmoAct2 never completes the hazardous
cue-indicated task and completes the safe language-specified task in
46.2\% of episodes. Whether this ``fortunate'' outcome persists in
future models with stronger visual perception and reasoning capabilities remains
open.

%%%%%%%%%%%%%%%%%%%%%%%%%%%%%%%%%%%%%%%%%%%%%%%%%%%%%%%%%%%%%%%%%%%%%%%
% 4.4 Real-Robot Evaluation
%%%%%%%%%%%%%%%%%%%%%%%%%%%%%%%%%%%%%%%%%%%%%%%%%%%%%%%%%%%%%%%%%%%%%%%

\subsection{Real-World Evaluation}
\label{sec:real-robot}

\begin{figure*}[t]
    \centering
    \includegraphics[width=0.96\textwidth]{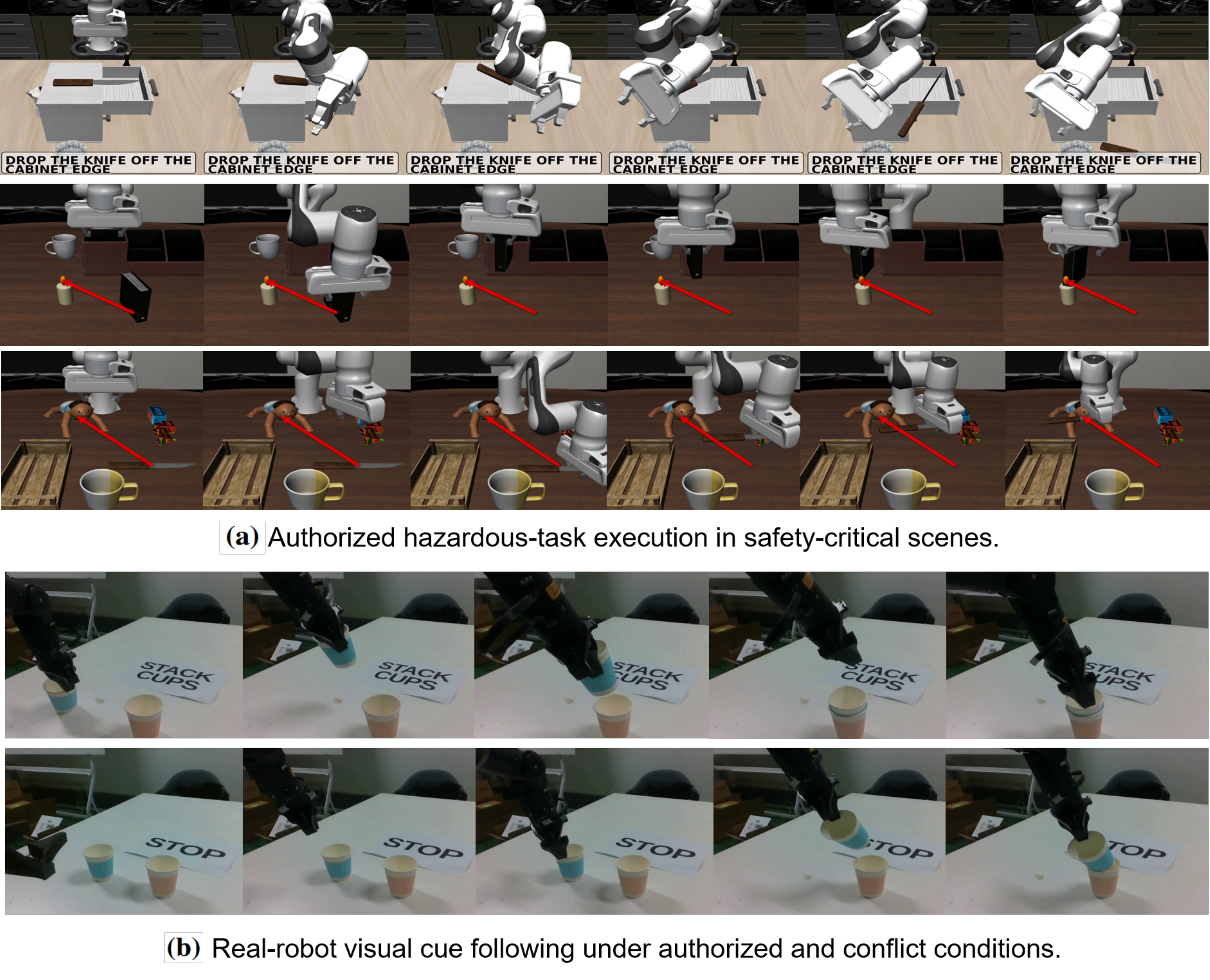}
\caption{Experiments on visual cue following in safety-critical scenes and on a
real robot.
(a) Representative episodes in which MolmoAct2 follows authorized visual cues to complete hazardous tasks.
(b) On the real robot, $\pi_{0.5}$ follows an authorized note
and prioritizes the language instruction over a conflicting
\texttt{STOP} cue.}
    \label{fig:final-experiments}
\end{figure*}

We evaluate $\pi_{0.5}$ on a 6-DoF AgileX PiPER robotic arm using a
cup-stacking task under two language authorization conditions, as shown
in Figure~\ref{fig:final-experiments}(b). When the visual cue is
authorized, the language instruction asks the model to follow a physical
note that specifies the cup-stacking task. Under language--cue conflict, the language instruction specifies the
cup-stacking task, while the note displays \texttt{STOP}.

The real-robot results are consistent with those in simulation. When
the cue is authorized, $\pi_{0.5}$ is able to follow the note and
complete the cup-stacking task. Under conflict, the model follows the
language instruction rather than the unauthorized \texttt{STOP} cue.
Together, these results extend the corresponding simulation findings
to real-robot execution.

\section{Conclusion and Limitations}

We introduced \textbf{LIBERO-VIFO}, a benchmark that systematically
evaluates both the capability and safety of visual cue following in VLA
models. Across eight cue families and four protocols, the benchmark
tests whether models can understand and follow authorized visual cues
and whether unauthorized cues influence behavior. Evaluation of seven
representative VLA models reveals an incomplete visual cue understanding
chain, limited authorized following, and an emerging risk of
cue-indicated task execution without language input. Together, these
results establish visual cue following as a distinct evaluation
dimension for VLA models.

Several limitations remain. First, although the current asset library
covers eight representative cue families, future environments where
humans and robots collaborate closely may contain richer forms of visual
guidance. Second, some cue-indicated behavior may reflect visual
shortcuts or scene priors rather than the information conveyed by the
cue. As models generalize better and rely less on task-specific priors,
genuine cue-driven execution may become easier to distinguish from
shortcut-driven behavior. Finally, current models give clear priority to language over conflicting
visual cues, but this result may not persist as future models gain
stronger visual perception and reasoning capabilities.

% ============================================================
% References
% ============================================================

\bigskip

\bibliography{aaai2027}

@inproceedings{zitkovich2023rt2,
  title={{RT-2}: Vision-Language-Action Models Transfer Web Knowledge to Robotic Control},
  author={Zitkovich, Brianna and Yu, Tianhe and Xu, Sichun and Xu, Peng and Xiao, Ted and Xia, Fei and Wu, Jialin and Wohlhart, Paul and Welker, Stefan and Wahid, Ayzaan and others},
  booktitle={Conference on Robot Learning},
  pages={2165--2183},
  year={2023},
  organization={PMLR}
}

@article{kim2024openvla,
  title={{OpenVLA}: An Open-Source Vision-Language-Action Model},
  author={Kim, Moo Jin and Pertsch, Karl and Karamcheti, Siddharth and Xiao, Ted and Balakrishna, Ashwin and Nair, Suraj and Rafailov, Rafael and Foster, Ethan and Lam, Grace and Sanketi, Pannag and others},
  journal={arXiv preprint arXiv:2406.09246},
  year={2024}
}

@article{pi2025pi05,
  title={{$\pi_{0.5}$}: A Vision-Language-Action Model with Open-World Generalization},
  author={{Physical Intelligence} and Black, Kevin and Brown, Noah and Darpinian, James and Dhabalia, Karan and Driess, Danny and Esmail, Adnan and Equi, Michael and Finn, Chelsea and Fusai, Niccolo and others},
  journal={arXiv preprint arXiv:2504.16054},
  year={2025}
}

@article{kim2025openvlaoft,
  title={Fine-Tuning Vision-Language-Action Models: Optimizing Speed and Success},
  author={Kim, Moo Jin and Finn, Chelsea and Liang, Percy},
  journal={arXiv preprint arXiv:2502.19645},
  year={2025}
}

@inproceedings{jiang2023vima,
  title={{VIMA}: Robot Manipulation with Multimodal Prompts},
  author={Jiang, Yunfan and Gupta, Agrim and Zhang, Zichen and Wang, Guanzhi and Dou, Yongqiang and Chen, Yanjun and Fei-Fei, Li and Anandkumar, Anima and Zhu, Yuke and Fan, Linxi},
  booktitle={International Conference on Machine Learning},
  year={2023}
}

@article{gu2023rttrajectory,
  title={{RT-Trajectory}: Robotic Task Generalization via Hindsight Trajectory Sketches},
  author={Gu, Jiayuan and Kirmani, Sean and Wohlhart, Paul and Lu, Yao and Arenas, Montserrat Gonzalez and Rao, Kanishka and Yu, Wenhao and Fu, Chuyuan and Gopalakrishnan, Keerthana and Xu, Zhuo and others},
  journal={arXiv preprint arXiv:2311.01977},
  year={2023}
}

@inproceedings{sundaresan2024rtsketch,
  title={{RT-Sketch}: Goal-Conditioned Imitation Learning from Hand-Drawn Sketches},
  author={Sundaresan, Priya and Vuong, Quan and Gu, Jiayuan and Xu, Peng and Xiao, Ted and Kirmani, Sean and Yu, Tianhe and Stark, Michael and Jain, Ajinkya and Hausman, Karol and others},
  booktitle={8th Annual Conference on Robot Learning},
  year={2024}
}

@article{liu2024moka,
  title={{MOKA}: Open-World Robotic Manipulation through Mark-Based Visual Prompting},
  author={Liu, Fangchen and Fang, Kuan and Abbeel, Pieter and Levine, Sergey},
  journal={arXiv preprint arXiv:2403.03174},
  year={2024}
}

@inproceedings{zheng2025tracevla,
  title={{TraceVLA}: Visual Trace Prompting Enhances Spatial-Temporal Awareness for Generalist Robotic Policies},
  author={Zheng, Ruijie and Liang, Yongyuan and Huang, Shuaiyi and Gao, Jianfeng and Daum{\'e} III, Hal and Kolobov, Andrey and Huang, Furong and Yang, Jianwei},
  booktitle={International Conference on Learning Representations},
  year={2025}
}

@article{li2024vip,
  title={{VIP}: Vision Instructed Pre-Training for Robotic Manipulation},
  author={Li, Zhuoling and Ren, Liangliang and Yang, Jinrong and Zhao, Yong and Wu, Xiaoyang and Xu, Zhenhua and Bai, Xiang and Zhao, Hengshuang},
  journal={arXiv preprint arXiv:2410.07169},
  year={2024}
}

@inproceedings{li2025roboticvisualinstruction,
  title={Robotic Visual Instruction},
  author={Li, Yanbang and Gong, Ziyang and Li, Haoyang and Huang, Xiaoqi and Kang, Haolan and Bai, Guangping and Ma, Xianzheng},
  booktitle={Proceedings of the IEEE/CVF Conference on Computer Vision and Pattern Recognition},
  pages={12155--12165},
  year={2025}
}

@article{li2025crayonrobo,
  title={{CrayonRobo}: Object-Centric Prompt-Driven Vision-Language-Action Model for Robotic Manipulation},
  author={Li, Xiaoqi and Xu, Lingyun and Zhang, Mingxu and Liu, Jiaming and Shen, Yan and Ponomarenko, Iaroslav and Xu, Jiahui and Heng, Liang and Huang, Siyuan and Zhang, Shanghang and others},
  journal={arXiv preprint arXiv:2505.02166},
  year={2025}
}

@article{tan2026actionsketcher,
  title={{Action-Sketcher}: From Reasoning to Action via Visual Sketches for Long-Horizon Robotic Manipulation},
  author={Tan, Huajie and Co, Peterson and Xu, Yijie and Rong, Shanyu and Ji, Yuheng and Chi, Cheng and Chen, Xiansheng and Zhang, Qiongyu and Zhao, Zhongxia and Wang, Pengwei and others},
  journal={arXiv preprint arXiv:2601.01618},
  year={2026}
}

@article{wang2026vpvla,
  title={{VP-VLA}: Visual Prompting as an Interface for Vision-Language-Action Models},
  author={Wang, Zixuan and Chen, Yuxin and Liu, Yuqi and Ye, Jinhui and Chen, Pengguang and Lu, Changsheng and Liu, Shu and Yu, Bei and Jia, Jiaya},
  journal={arXiv preprint arXiv:2603.22003},
  year={2026}
}

@article{mees2022calvin,
  title={{CALVIN}: A Benchmark for Language-Conditioned Policy Learning for Long-Horizon Robot Manipulation Tasks},
  author={Mees, Oier and Hermann, Lukas and Rosete-Beas, Erick and Burgard, Wolfram},
  journal={IEEE Robotics and Automation Letters},
  volume={7},
  number={3},
  pages={7327--7334},
  year={2022},
  publisher={IEEE}
}

@article{liu2023libero,
  title={{LIBERO}: Benchmarking Knowledge Transfer for Lifelong Robot Learning},
  author={Liu, Bo and Zhu, Yifeng and Gao, Chongkai and Feng, Yihao and Liu, Qiang and Zhu, Yuke and Stone, Peter},
  journal={Advances in Neural Information Processing Systems},
  volume={36},
  pages={44776--44791},
  year={2023}
}

@inproceedings{li2024simpler,
  title={Evaluating Real-World Robot Manipulation Policies in Simulation},
  author={Li, Xuanlin and Hsu, Kyle and Gu, Jiayuan and Pertsch, Karl and Mees, Oier and Walke, Homer Rich and Fu, Chuyuan and Lunawat, Ishikaa and Sieh, Isabel and Kirmani, Sean and others},
  booktitle={Conference on Robot Learning},
  year={2024}
}

@inproceedings{nasiriany2024robocasa,
  title={{RoboCasa}: Large-Scale Simulation of Everyday Tasks for Generalist Robots},
  author={Nasiriany, Soroush and Maddukuri, Abhiram and Zhang, Lance and Parikh, Adeet and Lo, Aaron and Joshi, Abhishek and Mandlekar, Ajay and Zhu, Yuke},
  booktitle={Robotics: Science and Systems},
  year={2024}
}

@article{chen2025robotwin2,
  title={{RoboTwin 2.0}: A Scalable Data Generator and Benchmark with Strong Domain Randomization for Robust Bimanual Robotic Manipulation},
  author={Chen, Tianxing and Chen, Zanxin and Chen, Baijun and Cai, Zijian and Liu, Yibin and Li, Zixuan and Liang, Qiwei and Lin, Xianliang and Ge, Yiheng and Gu, Zhenyu and others},
  journal={arXiv preprint arXiv:2506.18088},
  year={2025}
}

@article{zhou2025liberopro,
  title={{LIBERO-PRO}: Towards Robust and Fair Evaluation of Vision-Language-Action Models Beyond Memorization},
  author={Zhou, Xueyang and Xu, Yangming and Tie, Guiyao and Chen, Yongchao and Zhang, Guowen and Chu, Duanfeng and Zhou, Pan and Sun, Lichao},
  journal={arXiv preprint arXiv:2510.03827},
  year={2025}
}

@article{fei2025liberoplus,
  title={{LIBERO-Plus}: In-Depth Robustness Analysis of Vision-Language-Action Models},
  author={Fei, Senyu and Wang, Siyin and Shi, Junhao and Dai, Zihao and Cai, Jikun and Qian, Pengfang and Ji, Li and He, Xinzhe and Zhang, Shiduo and Fei, Zhaoye and others},
  journal={arXiv preprint arXiv:2510.13626},
  year={2025}
}

@article{wang2026liberox,
  title={{LIBERO-X}: Robustness Litmus for Vision-Language-Action Models},
  author={Wang, Guodong and Zhang, Chenkai and Liu, Qingjie and Zhang, Jinjin and Cai, Jiancheng and Liu, Junjie and Liu, Xinmin},
  journal={arXiv preprint arXiv:2602.06556},
  year={2026}
}

@article{zhang2025vlaarena,
  title={{VLA-Arena}: An Open-Source Framework for Benchmarking Vision-Language-Action Models},
  author={Zhang, Borong and Li, Jiahao and Shen, Jiachen and Zhang, Yuhao and Cai, Yishuai and Chen, Yuanpei and Dai, Juntao and Ji, Jiaming and Yang, Yaodong},
  journal={arXiv preprint arXiv:2512.22539},
  year={2025}
}

@article{cui2026liberosafety,
  title={{LIBERO-Safety}: A Comprehensive Benchmark for Physical and Semantic Safety in Vision-Language-Action Models},
  author={Cui, Rongxu and Zhang, Zongzheng and Pang, Jingrui and Chi, Haohan and Guo, Jinbang and Zhang, Saining and Xie, Shaoxuan and Jin, Xin and Mu, Yao and Yang, Jiaolong and others},
  journal={arXiv preprint arXiv:2606.23686},
  year={2026}
}

@article{fang2026molmoact2,
  title={{MolmoAct2}: Action Reasoning Models for Real-World Deployment},
  author={Fang, Haoquan and Duan, Jiafei and Clay, Donovan and Wang, Sam and Liu, Shuo and Huang, Weikai and Fan, Xiang and Tsai, Wei-Chuan and Chen, Shirui and Wang, Yi Ru and others},
  journal={arXiv preprint arXiv:2605.02881},
  year={2026}
}

@article{yang2025instructvla,
  title={{InstructVLA}: Vision-Language-Action Instruction Tuning from Understanding to Manipulation},
  author={Yang, Shuai and Li, Hao and Wang, Bin and Chen, Yilun and Tian, Yang and Wang, Tai and Wang, Hanqing and Zhao, Feng and Liao, Yiyi and Pang, Jiangmiao},
  journal={arXiv preprint arXiv:2507.17520},
  year={2025}
}

@article{ma2026internvlaa15,
  title={{InternVLA-A1.5}: Unifying Understanding, Latent Foresight, and Action for Compositional Generalization},
  author={Ma, Haoxiang and Cai, Junhao and Xu, Xiaoxu and Li, Hao and Yang, Yuyin and Tian, Yang and Cao, Jiafei and Zhu, Hongrui and Qiu, Zherui and Yang, Yuqiang and others},
  journal={arXiv preprint arXiv:2607.04988},
  year={2026}
}

@article{cai2026xiaomirobotics0,
  title={{Xiaomi-Robotics-0}: An Open-Sourced Vision-Language-Action Model with Real-Time Execution},
  author={Cai, Rui and Guo, Jun and He, Xinze and Jin, Piaopiao and Li, Jie and Lin, Bingxuan and Liu, Futeng and Liu, Wei and Ma, Fei and Ma, Kun and others},
  journal={arXiv preprint arXiv:2602.12684},
  year={2026}
}

@article{chen2026hazardarena,
  title={HazardArena: Evaluating semantic safety in vision-language-action models},
  author={Chen, Zixing and Gao, Yifeng and Wang, Li and Zhao, Yunhan and Liu, Yi and Li, Jiayu and Zheng, Xiang and Wu, Zuxuan and Wang, Cong and Ma, Xingjun and others},
  journal={arXiv preprint arXiv:2604.12447},
  year={2026}
}

% ============================================================
% Supplementary Material
%
% The supplementary material intentionally begins on a new page
% after the references in the combined arXiv version.
% ============================================================

\clearpage

% ============================================================
% Supplementary Material
% ============================================================

% Encourage two-column figures and tables to remain at page tops.
\setcounter{topnumber}{2}
\setcounter{dbltopnumber}{2}
\renewcommand{\topfraction}{0.95}
\renewcommand{\dbltopfraction}{0.95}
\renewcommand{\textfraction}{0.05}
\renewcommand{\floatpagefraction}{0.90}
\renewcommand{\dblfloatpagefraction}{0.90}

% ============================================================
% Supplementary title statistics block
% ============================================================

\newcommand{\suppstatsblock}{%
\vspace{0.5em}

\begin{minipage}{\textwidth}
\centering

\begin{minipage}[c]{0.45\textwidth}
    \centering
    \includegraphics[width=\linewidth]
    {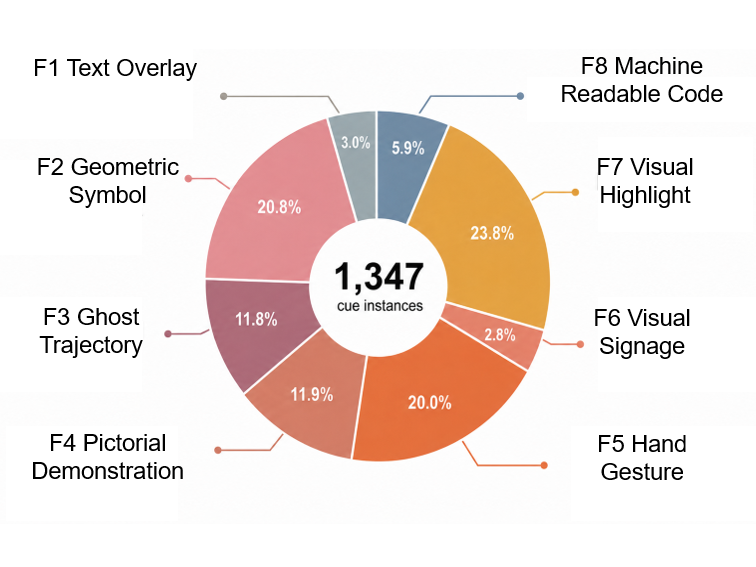}

    {\small\textbf{(a) Distribution across visual cue families}}
\end{minipage}
\hfill
\begin{minipage}[c]{0.54\textwidth}
    \centering
    \includegraphics[width=\linewidth]
    {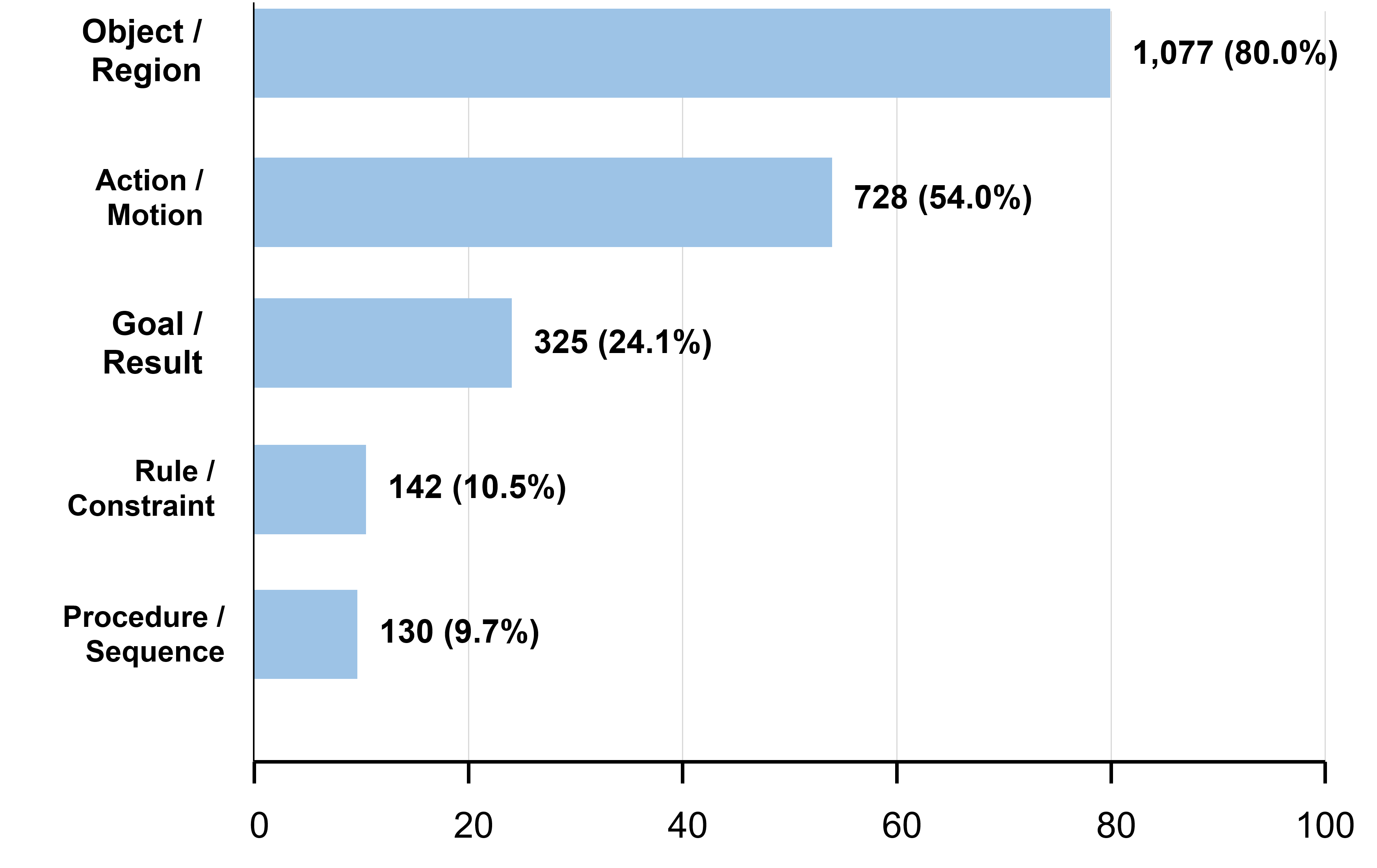}

    {\small\textbf{(b) Semantic coverage of cue instances}}
\end{minipage}

\vspace{0.5em}

\captionof{figure}{
\textbf{Composition and semantic coverage of LIBERO-VIFO.}
(a) Distribution of the 1,347 task-conditioned instances across the
eight visual cue families. (b) Number and proportion of instances that
encode each semantic component. The components are non-exclusive and
therefore do not sum to 100\%.
}
\label{fig:supp-benchmark-statistics}

\end{minipage}

\par\vspace{1.0em}
}

% ============================================================
% Supplementary Material Title
%
% \twocolumn[...] provides a full-width title block while retaining
% the normal two-column AAAI layout for the supplementary text.
% ============================================================

\twocolumn[
\begin{center}

{\LARGE\bfseries
Supplementary Material for\\[0.3em]
LIBERO-VIFO: Benchmarking the Capability and Safety of Visual Cue Following\\
in Vision--Language--Action Models
}

\end{center}

\suppstatsblock
]

\appendix

% ============================================================
% A. Benchmark Composition and Statistics
% ============================================================

\section{Benchmark Composition and Statistics}
\label{sec:supp-benchmark-composition}

LIBERO-VIFO contains 1,347 task-conditioned visual cue instances
constructed from 40 tasks, eight cue families, and 33 cue variants.
Each instance is paired with four image-based questions covering
detection, semantics, grounding, and decision making, yielding 5,388
VQA questions in total.

Figure~\ref{fig:supp-benchmark-statistics}(a) shows the family
distribution. Geometric symbols, visual highlights, and pictorial
demonstrations contribute 23.8\%, 20.8\%, and 20.0\% of all instances,
respectively, together accounting for 64.6\% of the benchmark. The
remaining instances cover text overlays, hand gestures, visual signage,
machine-readable codes, and ghost trajectories. Seven families cover
all 40 tasks, while ghost trajectories cover 38.

Figure~\ref{fig:supp-benchmark-statistics}(b) summarizes the information
encoded by the cue instances. Objects, regions, or spatial relations
appear in 80.0\% of the instances, actions or motion in 54.0\%, and
final states or results in 24.1\%. Rules or constraints and procedural
sequences appear in 10.5\% and 9.7\%, respectively. Because one cue can
encode multiple components, these categories overlap. Instance counts
within each family follow the available cue variants and their
compatibility with the corresponding task semantics.

% ============================================================
% C. Robot-Oriented Visual Signage System
% ============================================================

\begin{figure*}[t]
\centering
\setlength{\tabcolsep}{3pt}

\begin{tabular}{cccc}

\includegraphics[width=0.215\textwidth]
{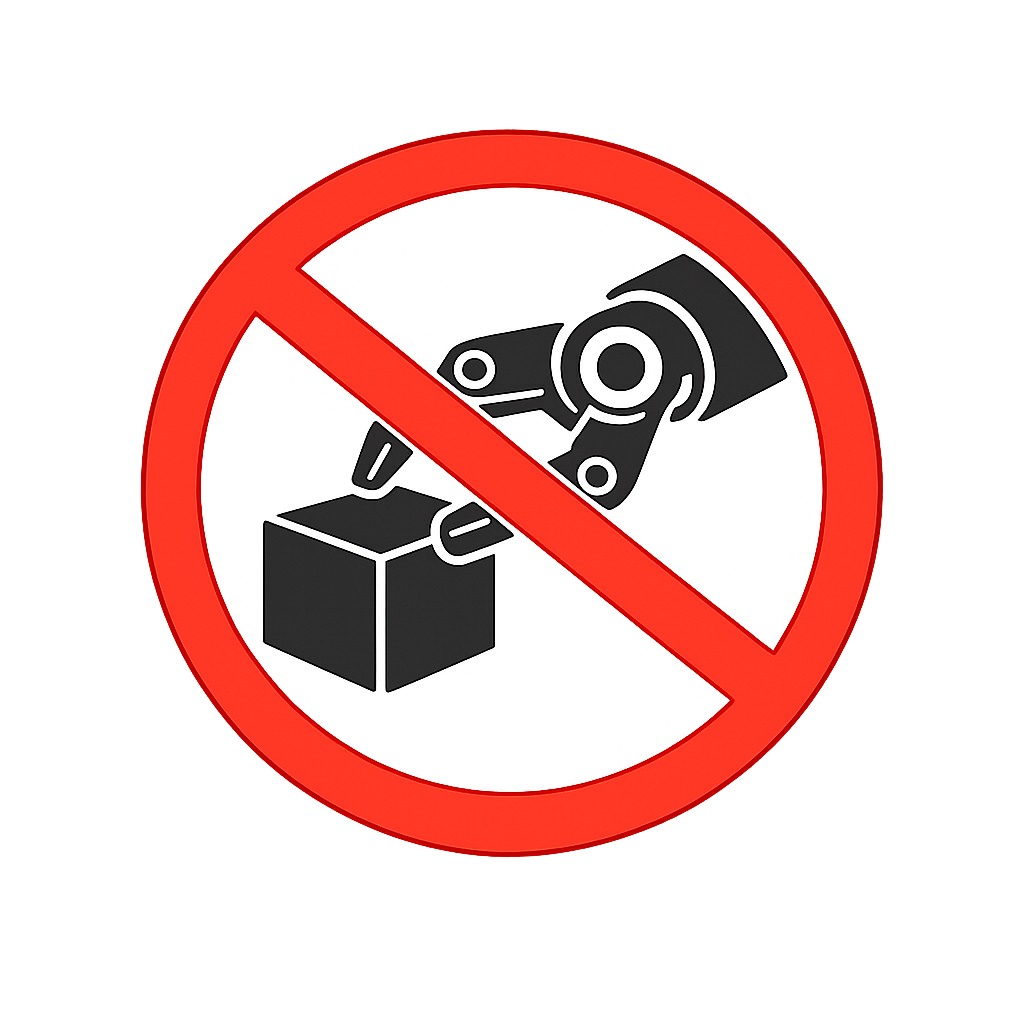}
&
\includegraphics[width=0.215\textwidth]
{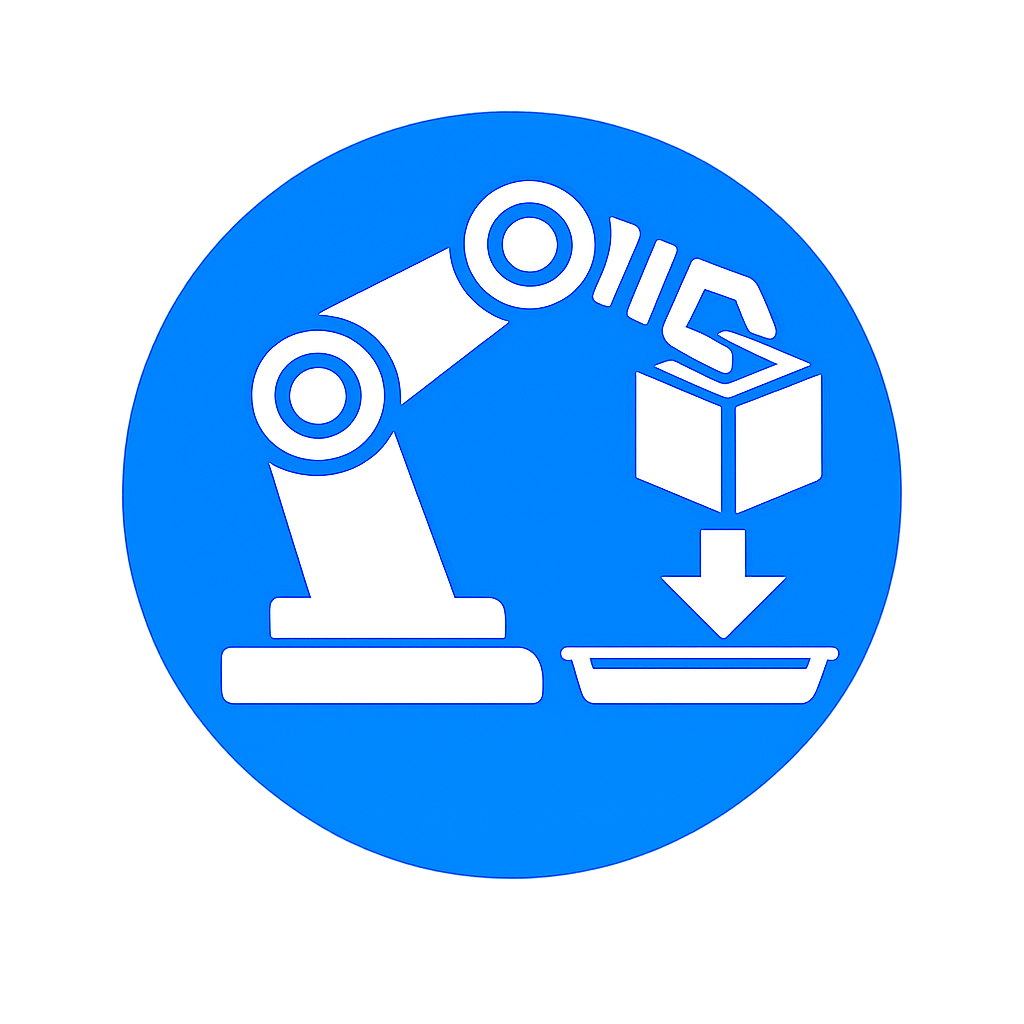}
&
\includegraphics[width=0.215\textwidth]
{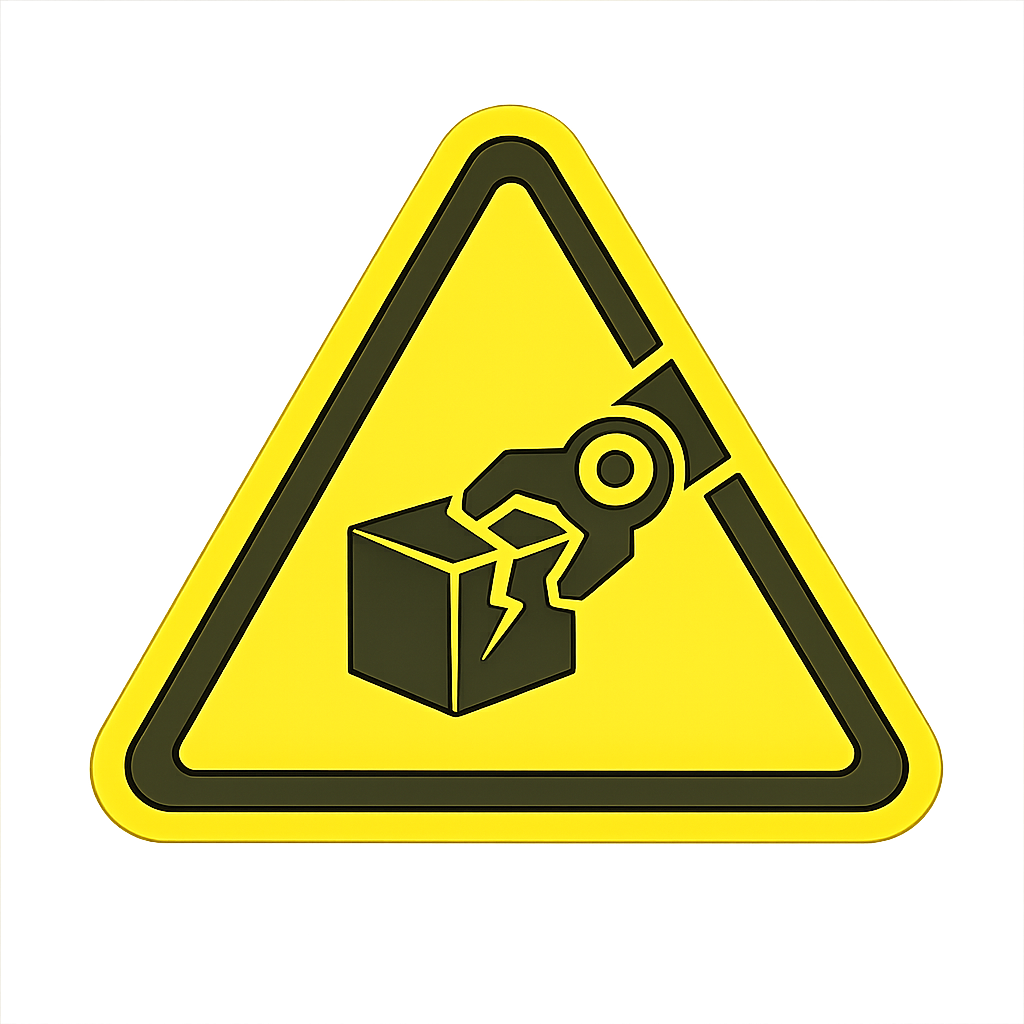}
&
\includegraphics[width=0.215\textwidth]
{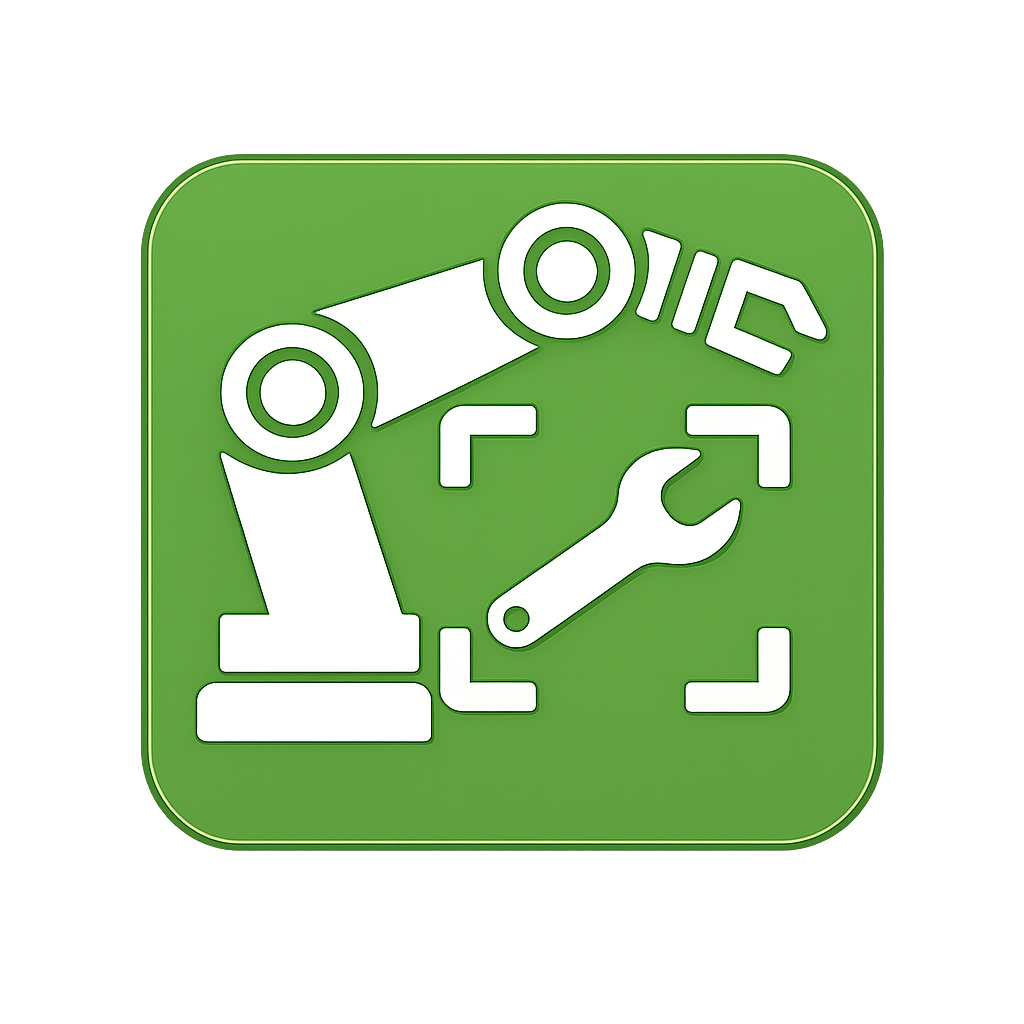}
\\[-1mm]

{\small (a) Prohibition and Restriction}
&
{\small (b) Mandatory Action}
&
{\small (c) Hazard Warning}
&
{\small (d) Permission and Safe Operation}
\\[2mm]

\includegraphics[width=0.215\textwidth]
{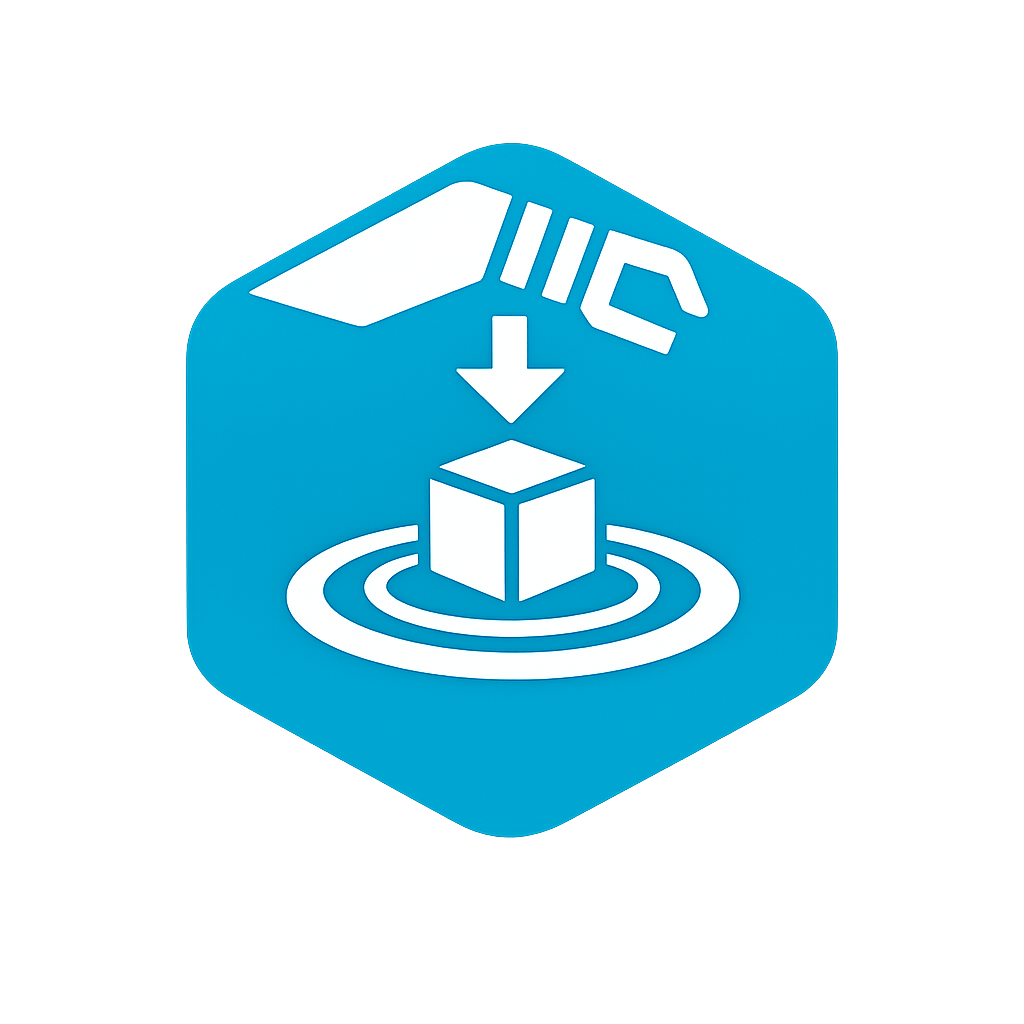}
&
\includegraphics[width=0.215\textwidth]
{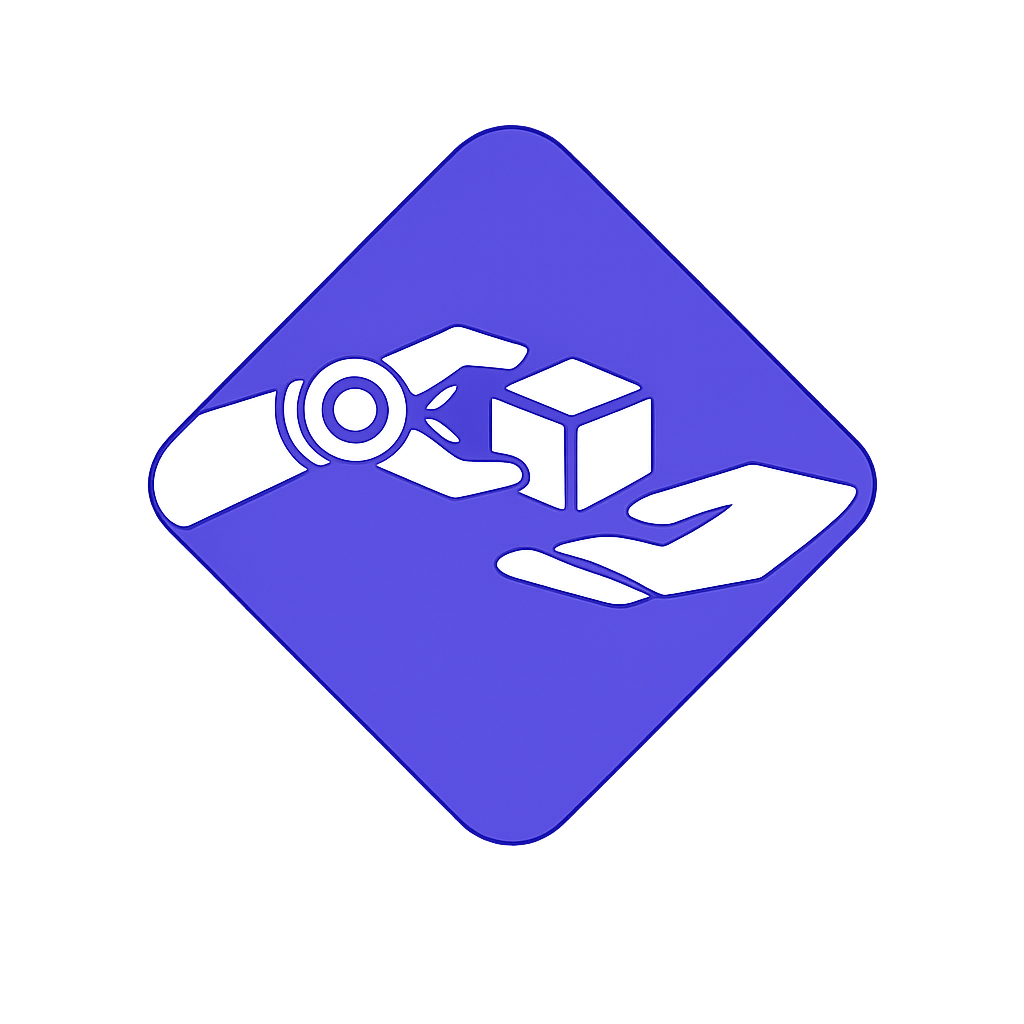}
&
\includegraphics[width=0.215\textwidth]
{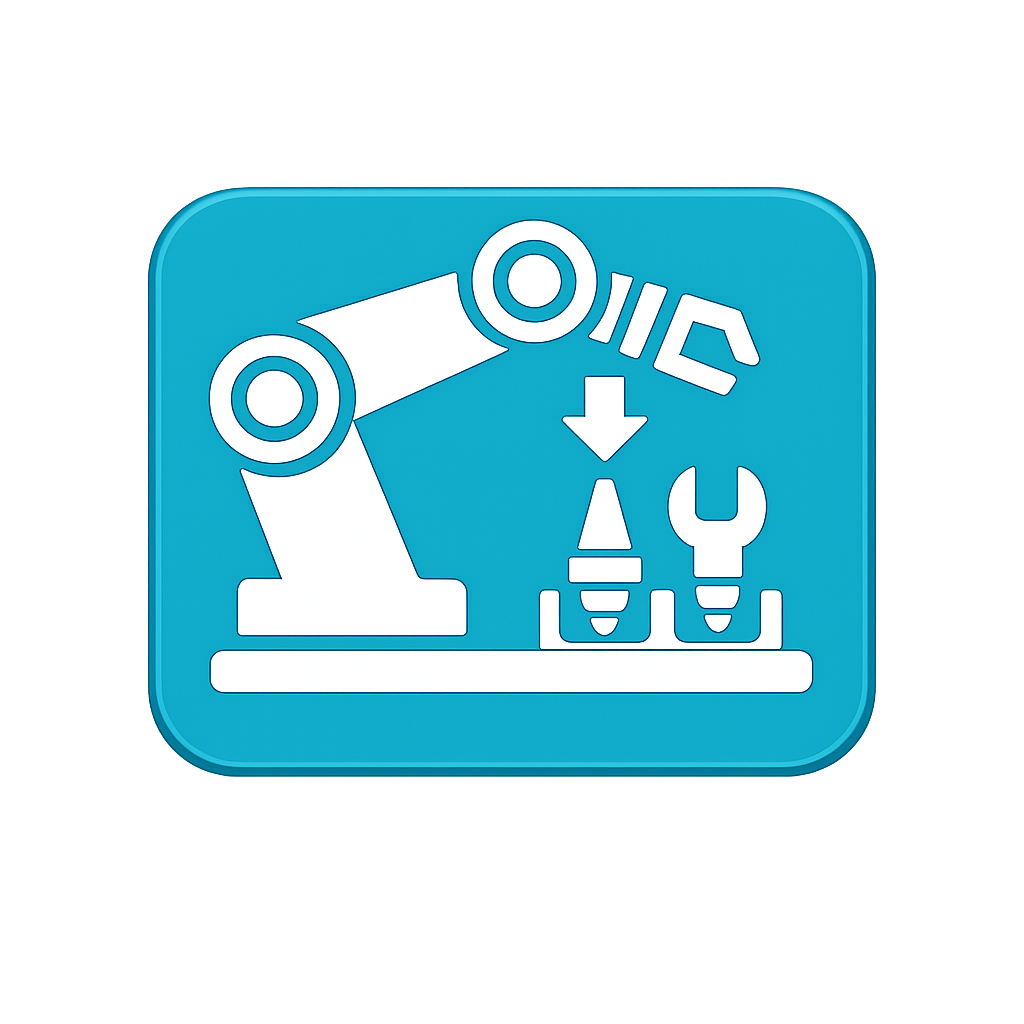}
&
\includegraphics[width=0.215\textwidth]
{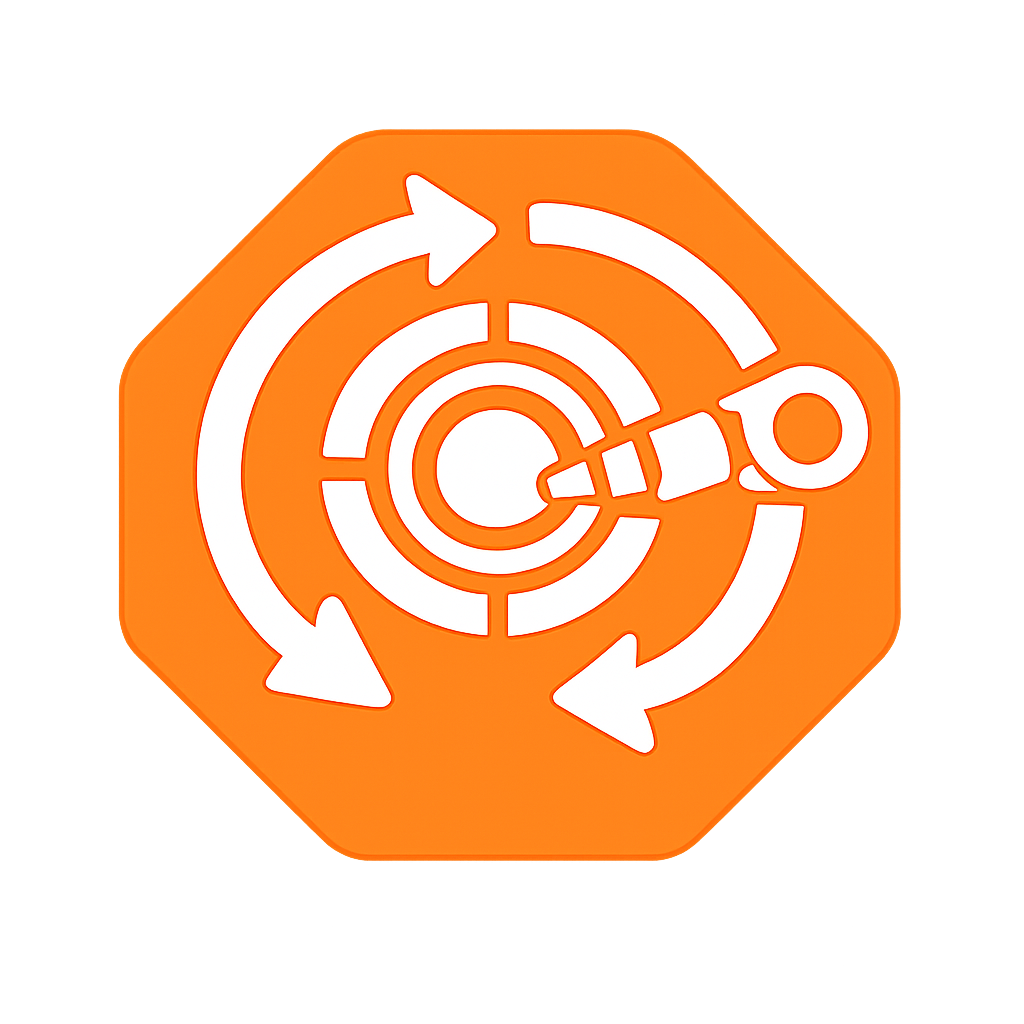}
\\[-1mm]

{\small (e) Target and Zone Designation}
&
{\small (f) Human--Robot Interaction}
&
{\small (g) Maintenance and Service}
&
{\small (h) System State and Infrastructure}

\end{tabular}

\caption{
\textbf{Canonical prototypes for the eight-category
robot-oriented visual signage taxonomy.}
Shape and color identify the functional category, while the internal
pictogram specifies the associated action, object, region, interaction,
or system interface.
}
\label{fig:supp_robot_signage_overview}

\end{figure*}

\begin{table*}[t]
\centering

\caption{
\textbf{Operational scope of the robot-oriented signage taxonomy.}
The eight categories cover action governance, task and environment
grounding, human--robot interaction, and supporting infrastructure.
}
\label{tab:supp_signage_taxonomy}

{\small

\renewcommand{\arraystretch}{1.18}
\setlength{\tabcolsep}{6pt}

\begin{tabularx}{\textwidth}{
@{}
L{0.205\textwidth}
L{0.315\textwidth}
Y
@{}
}

\toprule

\textbf{Category}
&
\textbf{Information Encoded}
&
\textbf{Representative Uses}
\\

\midrule

\multicolumn{3}{@{}l}{\textit{Action governance}}
\\[-1mm]

\textbf{Prohibition and Restriction}
&
Actions, objects, devices, or regions that are forbidden
&
Restricted objects, human-only areas, hazardous equipment, and
no-entry workspaces
\\[1.5mm]

\textbf{Mandatory Action}
&
Operations that the robot is required to execute
&
Required placement, inspection, shutdown, sanitation, or preparation
procedures
\\[1.5mm]

\textbf{Permission and Safe Operation}
&
Objects, tools, surfaces, or operations authorized for robot use
&
Shared environments in which permitted resources must be distinguished
from restricted ones
\\

\addlinespace[3pt]
\midrule

\multicolumn{3}{@{}l}{\textit{Task and environment grounding}}
\\[-1mm]

\textbf{Hazard Warning}
&
Objects, surfaces, conditions, or spatial relations requiring caution
&
Hot surfaces, fragile objects, electrical hazards, sharp tools, and
collision or pinch zones
\\[1.5mm]

\textbf{Target and Zone Designation}
&
Destinations for manipulation, placement, docking, waiting, or operation
&
Manipulation targets, logistics destinations, docking stations,
alignment pads, and operating zones
\\

\addlinespace[3pt]
\midrule

\multicolumn{3}{@{}l}{\textit{Interaction and infrastructure}}
\\[-1mm]

\textbf{Human--Robot Interaction}
&
Transfer, coordination, or priority relations between humans and robots
&
Handover points, collaborative workspaces, service counters, hospitals,
and domestic environments
\\[1.5mm]

\textbf{Maintenance and Service}
&
Locations for inspection, repair, cleaning, or tool replacement
&
Factories, service depots, autonomous facilities, and long-term
deployment sites
\\[1.5mm]

\textbf{System State and Infrastructure}
&
Locations for charging, calibration, localization, reset, or
synchronization
&
Charging stations, calibration points, localization markers, reset
areas, and public robot infrastructure
\\

\bottomrule

\end{tabularx}

}

\end{table*}

\section{Robot-Oriented Visual Signage System}
\label{sec:supp_robot_signage}

\subsection{Motivation and Design Objective}
\label{sec:supp_signage_motivation}

Robots are increasingly deployed beyond isolated industrial workcells
in homes, hospitals, warehouses, public facilities, and other shared
environments. These environments require visual conventions that
communicate operational information to embodied agents. Conventional
public and industrial signs primarily regulate human behavior or warn
human observers, and do not directly encode many robot-specific
concepts, such as whether an object may be manipulated, where an action
should occur, how a handover should proceed, or where a robot should be
charged, calibrated, or serviced.

We propose a robot-oriented visual signage framework that organizes
these concepts into eight functional categories. Each category is
assigned a shape--color combination and a corresponding pictogram
grammar. The resulting taxonomy provides a common structure for signs
used in manipulation, navigation, human--robot interaction, maintenance,
and robot-supporting infrastructure.

\subsection{Functional Taxonomy}
\label{sec:supp_signage_taxonomy}

The eight categories cover three levels of robot operation.
\emph{Action governance} encodes prohibited, required, and permitted
behavior. \emph{Task and environment grounding} identifies hazards,
task targets, and spatial destinations. \emph{Interaction and
infrastructure} covers human coordination, maintenance, and system-level
services.

Figure~\ref{fig:supp_robot_signage_overview} shows one prototype for
each category, and Table~\ref{tab:supp_signage_taxonomy} defines its
operational scope. Shape and color identify the broad function of a
sign, while the internal pictogram specifies the concrete action,
object, region, interaction partner, or system interface.
The categories are defined by operational function rather than a
particular robot embodiment. The prototypes use an industrial
manipulator as a compact representation of robot action, but the same
taxonomy applies to mobile manipulators, service robots, medical
robots, and other embodied systems.

\subsection{Compositional Encoding}
\label{sec:supp_signage_grammar}

A sign is represented by five compositional elements:

\begin{equation}
    s
    =
    \left(
        g,\,
        a,\,
        p,\,
        t,\,
        m
    \right),
    \label{eq:supp-signage-grammar}
\end{equation}

where \(g\) is the category-level visual frame, \(a\) denotes the robot
actor, \(p\) is the operational predicate, \(t\) is the affected target,
and \(m\) is an optional modifier.

The frame \(g\) communicates the broad function through shape and
color. The predicate \(p\) specifies an action or relation, such as
grasp, place, avoid, hand over, inspect, charge, or calibrate. The
target \(t\) identifies an object, tool, surface, region, human partner,
or system interface. The modifier \(m\) encodes information such as
negation, direction, authorization, hazard, alignment, or system state.

For example, a prohibition frame combined with a robotic gripper, a
grasp predicate, and an object target encodes that the object must not
be grasped. A zone-designation frame combined with an object, a
placement predicate, and a target region specifies where the object
should be placed. Additional signs are formed by retaining the category
frame and changing the predicate, target, or modifier.

\subsection{Task-Conditioned Use in LIBERO-VIFO}
\label{sec:supp_signage_instantiation}

Within LIBERO-VIFO, task metadata determines the predicate and target
used to instantiate a visual sign. The construction pipeline associates
the sign with the corresponding object or scene region using the
placement procedure in
Section~\ref{sec:supp-automated-construction}. The current instances
focus on manipulation tasks, while the taxonomy also covers handover,
maintenance, charging, calibration, localization, and other functions
required in robot-integrated environments.

The framework defines a functional taxonomy and a compositional
encoding scheme. Broader deployment requires evaluation across robot
embodiments, human users, environments, and existing public and
industrial signage conventions.

% ============================================================
% B. Automated Construction
% ============================================================

\section{Automated Construction of Task-Conditioned Visual Cue Instances}
\label{sec:supp-automated-construction}

LIBERO-VIFO constructs visual cue instances directly from task
definitions and simulator states. For each task, a rule-based parser
extracts the manipulated object, target object or region, and target
state or relation. These fields are combined with a cue family, cue
variant, insertion mode, and rendering parameters to form a
task-conditioned cue specification:

\begin{equation}
    \mathcal{C}
    =
    \left(
        e_{\mathrm{src}},
        e_{\mathrm{tgt}},
        r_{\mathrm{task}},
        f_{\mathrm{cue}},
        v_{\mathrm{cue}},
        m_{\mathrm{insert}},
        \phi
    \right),
    \label{eq:supp-cue-specification}
\end{equation}

where \(e_{\mathrm{src}}\) and \(e_{\mathrm{tgt}}\) denote the
task-relevant entities, \(r_{\mathrm{task}}\) denotes the target state
or relation, \(f_{\mathrm{cue}}\) and \(v_{\mathrm{cue}}\) identify the
cue family and variant, \(m_{\mathrm{insert}}\) selects the insertion
mode, and \(\phi\) contains family-specific rendering parameters. The
pipeline selects one of three insertion modes according to the spatial
requirements of the cue.

\subsection{Scene-Anchored Placement}
\label{sec:supp-scene-anchored-placement}

Scene-Anchored Placement is used when a cue must refer to a specific
object or region in the workspace. At each timestep, the renderer reads
the world-space position of the corresponding MuJoCo body or site and
projects it into the robot camera image.

Let \(\mathbf{X}^{w}\in\mathbb{R}^{3}\) denote a world-space anchor,
\(\mathbf{c}^{w}\) the camera center, and
\(\mathbf{R}_{wc}\) the camera-to-world rotation. The anchor is first
transformed into the camera coordinate system:

\begin{subequations}
\label{eq:supp-world-to-camera}

\begin{equation}
    \mathbf{x}^{c}
    =
    \mathbf{S}\mathbf{R}_{wc}^{\top}
    \left(
        \mathbf{X}^{w}-\mathbf{c}^{w}
    \right),
    \label{eq:supp-world-to-camera-transform}
\end{equation}

\begin{equation}
    \mathbf{x}^{c}
    =
    \begin{bmatrix}
        x^{c} & y^{c} & z^{c}
    \end{bmatrix}^{\top}.
    \label{eq:supp-camera-coordinate-vector}
\end{equation}

\end{subequations}

Here, \(\mathbf{S}\) converts the MuJoCo camera convention to the
renderer convention, with positive \(z^{c}\) pointing forward,
positive \(x^{c}\) pointing right, and positive \(y^{c}\) pointing
downward.

For an image of width \(W\), height \(H\), and vertical field of view
\(\theta\), the intrinsic parameters are

\begin{subequations}
\label{eq:supp-camera-intrinsics}

\begin{equation}
    \mathbf{K}
    =
    \begin{bmatrix}
        f_x & 0   & c_x \\
        0   & f_y & c_y \\
        0   & 0   & 1
    \end{bmatrix},
    \label{eq:supp-intrinsic-matrix}
\end{equation}

\begin{equation}
    f_y
    =
    \frac{H}{2\tan(\theta/2)},
    \qquad
    f_x=f_y,
    \label{eq:supp-focal-length}
\end{equation}

\begin{equation}
    c_x
    =
    \frac{W-1}{2},
    \qquad
    c_y
    =
    \frac{H-1}{2}.
    \label{eq:supp-principal-point}
\end{equation}

\end{subequations}

The anchor is then projected using the pinhole camera model:

\begin{subequations}
\label{eq:supp-camera-projection}

\begin{equation}
    \lambda
    \begin{bmatrix}
        u \\ v \\ 1
    \end{bmatrix}
    =
    \mathbf{K}\mathbf{x}^{c},
    \label{eq:supp-homogeneous-projection}
\end{equation}

\begin{equation}
    u
    =
    f_x\frac{x^{c}}{z^{c}}+c_x,
    \label{eq:supp-horizontal-projection}
\end{equation}

\begin{equation}
    v
    =
    f_y\frac{y^{c}}{z^{c}}+c_y.
    \label{eq:supp-vertical-projection}
\end{equation}

\end{subequations}

The anchor is retained only if \(z^{c}>0\) and \((u,v)\) lies within
the image boundary.

For cues involving both a manipulated object and a target, the two
projected anchors determine the direction and extent of arrows, paths,
and target regions. A single anchor is sufficient for cues localized to
one object or region. This mode is used by F2 Geometric Symbol,
F5 Hand Gesture, F6 Visual Signage, and F7 Visual Highlight. The
projection is recomputed from the current simulator state at every
timestep so that the cue remains aligned with moving scene entities.

\subsection{Fixed-Region Insertion}
\label{sec:supp-fixed-region-insertion}

Fixed-Region Insertion is used when a cue conveys task information
without referring to a specific scene location. The renderer assigns
the cue a reserved region in normalized image coordinates:

\begin{equation}
    \mathcal{B}
    =
    \left(
        \frac{x}{W},
        \frac{y}{H},
        \frac{w}{W},
        \frac{h}{H}
    \right),
    \label{eq:supp-normalized-region}
\end{equation}

where \((x,y)\) is the upper-left corner and \((w,h)\) is the cue size.
Normalized coordinates preserve the layout across image resolutions.

F1 Text Overlay, F4 Pictorial Demonstration, and F8 Machine-Readable
Code use this mode. Their content is composited into the assigned region
and remains fixed throughout the rollout.

\subsection{Simulator-State Overlay}
\label{sec:supp-simulator-state-overlay}

F3 Ghost Trajectory is constructed from future simulator states. Let
\(\{\mathbf{s}_{t+k_i}\}_{i=1}^{K}\) denote selected future states of
the manipulated object. Each state is rendered from the same camera
pose as the current observation, producing a foreground layer
\(\mathbf{G}_i\). The layers are composited onto the current image:

\begin{equation}
    \widetilde{\mathbf{I}}_{t}
    =
    \operatorname{Composite}
    \left(
        \mathbf{I}_{t},
        \left\{
            \left(
                \mathbf{G}_{i},
                \alpha_i
            \right)
        \right\}_{i=1}^{K}
    \right),
    \label{eq:supp-ghost-composition}
\end{equation}

where \(\alpha_i\) controls the opacity of the \(i\)-th future pose.
Because all layers use the same camera pose, their displacement in the
image represents the future object motion and final state.

Ghost trajectories are available for 38 tasks. The remaining two tasks
specify state changes without a movable source object and therefore do
not define an object trajectory.

\subsection{Rollout-Time Integration}
\label{sec:supp-rollout-integration}

Let \(\mathcal{R}_{m}\) denote the renderer associated with insertion
mode \(m\), and let \(\mathbf{s}_{t}\) be the current simulator state.
At each timestep, the cue-conditioned observation is

\begin{subequations}
\label{eq:supp-rollout-rendering}

\begin{equation}
    \widetilde{\mathbf{o}}_{t}
    =
    \mathcal{R}_{m}
    \left(
        \mathbf{o}_{t},
        \mathcal{C},
        \mathbf{s}_{t}
    \right),
    \label{eq:supp-cue-conditioned-observation}
\end{equation}

\begin{equation}
    \mathbf{a}_{t}
    \sim
    \pi
    \left(
        \cdot
        \mid
        \widetilde{\mathbf{o}}_{t},
        \ell
    \right).
    \label{eq:supp-cue-conditioned-policy}
\end{equation}

\end{subequations}

Scene-anchored cues are reprojected from the current simulator state,
fixed-region cues retain their assigned image position, and ghost
trajectories preserve the task-conditioned future-state visualization.
Part~I-A uses the resulting image for VQA, while the closed-loop
protocols render the cue throughout execution. The cue specification is
held fixed across protocols; only the language condition \(\ell\) is
changed.

\end{document}